\documentclass[a4paper,conference]{IEEEtran}
\usepackage{amsmath,amssymb} 
\usepackage{booktabs}
\usepackage{tabularx}
\usepackage{multirow}

\usepackage{cite}

\usepackage{cleveref}
\Crefformat{figure}{#2Fig.~#1#3}
\Crefmultiformat{figure}{Figs.~#2#1#3}{ and~#2#1#3}{, #2#1#3}{ and~#2#1#3}

\usepackage{bm}
\usepackage{enumitem}

\newcommand{\vpara}[1]{\vspace{0.05in}\noindent\textbf{#1}}

\usepackage{tikz}
\usetikzlibrary{shapes.geometric}

\usepackage{color}
\definecolor{Orange}{rgb}{0.9,0.5,0}
\definecolor{NavyBlue}{rgb}{0.1, 0.4, 0.8}
\definecolor{Magenta}{rgb}{0.8, 0.1, 0.6}
\definecolor{mypink2}{RGB}{219, 48, 122}

\usepackage{amsmath,amssymb} 

\begin{document}
%
\title{Semantic-Guided Inpainting Network for Complex Urban Scenes Manipulation}

\author{\IEEEauthorblockN{Pierfrancesco Ardino\IEEEauthorrefmark{1}\IEEEauthorrefmark{2}, Yahui Liu\IEEEauthorrefmark{1}\IEEEauthorrefmark{2},
Elisa Ricci\IEEEauthorrefmark{1}\IEEEauthorrefmark{2}, Bruno Lepri\IEEEauthorrefmark{1} and Marco De Nadai\IEEEauthorrefmark{1}}
\IEEEauthorblockA{\IEEEauthorrefmark{1}FBK,
Trento, Italy}
\IEEEauthorblockA{\IEEEauthorrefmark{2}University of Trento,
Trento, Italy\\
Email: \IEEEauthorrefmark{1}[surname]@fbk.eu}
}


%


\maketitle

\begin{abstract}
Manipulating images of complex scenes to reconstruct, insert and/or remove specific object instances is a challenging task.
Complex scenes contain multiple semantics and objects, which are frequently cluttered or ambiguous, thus hampering the performance of inpainting models.
Conventional techniques often rely on structural information such as object contours in multi-stage approaches that generate unreliable results and boundaries.
In this work, we propose a novel deep learning model to alter a complex urban scene by removing a user-specified portion of the image and coherently inserting a new object (e.g. car or pedestrian) in that scene.
Inspired by recent works on image inpainting, our proposed method leverages the semantic segmentation to model the content and structure of the image, and learn the best shape and location of the object to insert.
To generate reliable results, we design a new decoder block that combines the semantic segmentation and generation task to guide better the generation of new objects and scenes, which have to be semantically consistent with the image.
Our experiments, conducted on two large-scale datasets of urban scenes (Cityscapes and Indian Driving), show that our proposed approach successfully address the problem of semantically-guided inpainting of complex urban scene.
\end{abstract}


%
\IEEEpeerreviewmaketitle

\section{Introduction}

Manipulating images to insert and remove objects automatically is of paramount relevance for a large number of real-world applications including data augmentation, photo editing and Augmented Reality (AR).
Recent literature has shown promising results in image manipulation to translate images from one domain to another~\cite{isola2017image, zhu2017unpaired, choi2019stargan, liu2020gmm,siarohin2019animating,siarohin2019first, 10.1145/3394171.3413505}, inpaint missing parts of images~\cite{yu2019region, nazeri2019edgeconnect, jo2019sc}, and change portion of faces and fashion garments~\cite{han2019finet, jo2019sc}.
However, manipulating images (i.e. adding, reconstructing and removing objects) of complex scenes is a challenging and still unsolved problem.
Complex scenes are indeed characterized by multiple semantics and objects, which are often cluttered or occluded, making it difficult to apply conventional image inpainting techniques.

Existing works on image editing focus either on object insertion or removal. Solutions to the former task usually require users to draw segmentation pixels~\cite{hong2018learning} or a precise bounding box of the object to be inserted~\cite{lee2018context}, while object removal has been addressed by image inpainting~\cite{yu2019region, nazeri2019edgeconnect, jo2019sc}, which reconstructs the most probable pattern from contextual pixels. 
Thus, one has to choose between models that insert objects in controlled settings and models that inpaint corrupted areas.
Moreover, complex scenes (e.g. urban scenes) are often overlooked in favour of natural scenes and photographs with a small number of semantic classes, which allows the use of different stratagems to guide the generative network. In non-complex scenes, literature relies on edge maps~\cite{nazeri2019edgeconnect} and object contours~\cite{xiong2019foreground}, which results are however not satisfactory when the missing region is large or complex. 

\begin{figure}[t!]
   \centering
   \includegraphics[width=\linewidth]{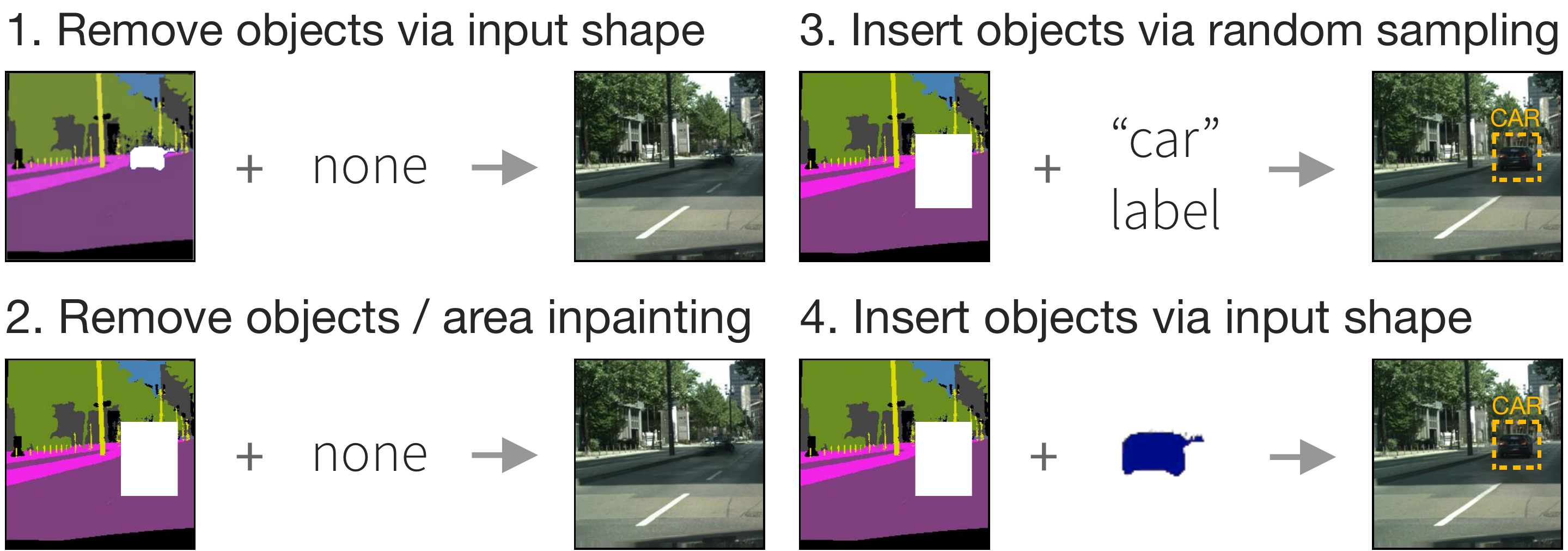}
   \caption{Our holistic model can be applied in a wide range of manipulations in complex scenes. At inference time users can remove objects by either precisely indicating a mask or an entire area. Moreover, they can insert new objects (e.g. cars, pedestrians) by randomly generating them or by feeding a segmentation mask as input.}
   \label{fig:teaser}
\end{figure}

In this paper, we focus on complex urban scenes that contain multiple objects, clutter and numerous semantic classes.
We propose a novel and unified framework to manipulate images by removing and inserting objects. We formulate the problem by learning to reconstruct (inpaint) missing regions and generate plausible shapes of objects to be inserted. 
To help the model at understanding complex scenes, we leverage information from semantic segmentation maps in order to guide the network in both the encoding and the decoding phases.
We learn to accurately generate both the semantic segmentation and the real pixels enforcing them to be consistent with each other. To this end, we design and propose a novel decoder module based on Spatially-Adaptive (DE) normalization (SPADE)~\cite{park2019semantic} that uses the predicted segmentation to normalize generated features and synthesize high-quality images.
Differently from previous works, our unified framework allows numerous use cases at inference time. For example, users can interact with the model by precisely removing the pixel of an object, removing entire areas, inserting random sampled objects or placing input shapes directly in the scene (see \Cref{fig:teaser}).

To evaluate our framework, we conduct experiments on two large-scale datasets of urban scenes, namely Cityscapes~\cite{cordts2016cityscapes} and Indian Driving~\cite{varma2019idd}. These datasets consist of a large number of semantic classes (on average 17) and diverse and unstructured environmental conditions, which make them a perfect benchmark for complex scene manipulations.
Our results show that with the proposed method we can insert and place objects in existing complex scenes generating high-quality images, outperforming state-of-the-art approaches on image inpainting.

Overall, the main contributions of our work are as follows: 
\begin{itemize}
    \item We propose a new holistic framework to manipulate complex scenes and allow users to insert and remove different types of objects in a single pass. At inference time, users can do a wide range of manipulations, considering different types of object insertion and removal operations.
    \item We design a new decoder module based on SPADE~\cite{park2019semantic} that uses the predicted segmentation map instead of the ground truth. Thus, at inference time we allow the use of SPADE without asking the user to specify the precise semantic pixels of the desired transformation. 
    \item We validate the proposed solution in the challenging task of manipulation of urban scenes using two large-scale datasets, namely Cityscapes and Indian Driving. Quantitative and qualitative results show that our method significantly outperforms the state of the art models in all the experiments.
\end{itemize}

\section{Related Work}
Our work is best placed in the literature of image inpainting and image manipulation. The former aims at restoring a damaged image or remove undesired objects, while the latter tries to synthetize new images with a user-specified object. 

\vpara{Image inpaiting}. Image inpainting approaches have witnessed a dramatic improvement in image quality especially thanks to deep learning methods, in particular to Generative Adversarial Networks (GANs)~\cite{goodfellow2014generative}. Most notably, Pathak  \textit{et al.}~\cite{pathak2016context} propose an encoder-decoder network, inspired by auto-encoder approaches, which synthetizes a part region of the image depending on its surroundings.
The combination of reconstruction and adversarial losses are shown to be sufficient for the neural networks to inpaint the missing region of the image. However, blurry results and disconnected edges frequently occur in generated regions. 
Thus, Iizuka \textit{et al.}~\cite{iizuka2017globally} propose to use dilated convolutions to increase the receptive field and combine the global and local discriminators to improve the quality of the generated patches. 
Yu \textit{et al.}~\cite{yu2018generative} instead propose the use of a two-stage approach to first coarsely reconstruct the image and then refine the coarse details. An additional contextual attention module is proposed to capture distant information. 
Zheng \textit{et al.}~\cite{zheng2019pluralistic} focus on generating multiple plausible results for the same missing region by using a Variational-Autoencoder (VAE).
Finally, Yu \textit{et al.}~\cite{yu2019region} propose a spatial region-wise approach that normalizes the corrupted and uncorrupted regions with two different means and variances.

Recently, various efforts using structural information have been explored to better reconstruct edges and contours.
For example, Nazeri \textit{et al.} \cite{nazeri2019edgeconnect} and Jo \textit{et al.} \cite{jo2019sc} use input edge maps during the image inpainting. Xiong \textit{et al.}~\cite{xiong2019foreground} use object contours and a multi-stage process to disentangle the background from a foreground object.
Ren \textit{et al.}~\cite{ren2019structureflow} instead focus on reconstructing missing structures of free-form missing parts through a flow-based method. 
Song \textit{et al.}~\cite{song2018spg} explore semantic segmentation images to guide the reconstruction in a two-stage process.
However, most of the techniques found to be effective in non-complex scenes (e.g. faces, simple scenes) result in non-satisfactory reconstructions due to the higher clutter and number of different semantics of complex scenes. 
Moreover, existing approaches based on image inpainting remove existing objects and reconstruct the background patterns without inserting new objects. 

\vpara{Image manipulation}. Image manipulation aims at modifying an existing image towards a user-desired outcome. In image-to-image translation this outcome is usually changing the visual appearance of an image while maintaining the original structure (e.g. blonde hair woman $\leftrightarrow$ black hair man). For example, literature has shown promising results on translating domains in the image space (e.g. blonde hair woman to black hair woman)~\cite{isola2017image, zhu2017unpaired, choi2019stargan, liu2020gmm}, and from the image space to the semantic space~\cite{isola2017image, wang2018high}. 
There has been also some efforts towards free-form editing, where the mask is not rectangular nor regular. Notable example is free-form editing of faces and fashion pictures where people draw a sketch of the desired transformation~\cite{dong2019fashion, jo2019sc}.

\begin{figure*}[ht]
   \centering
   \includegraphics[width=\linewidth]{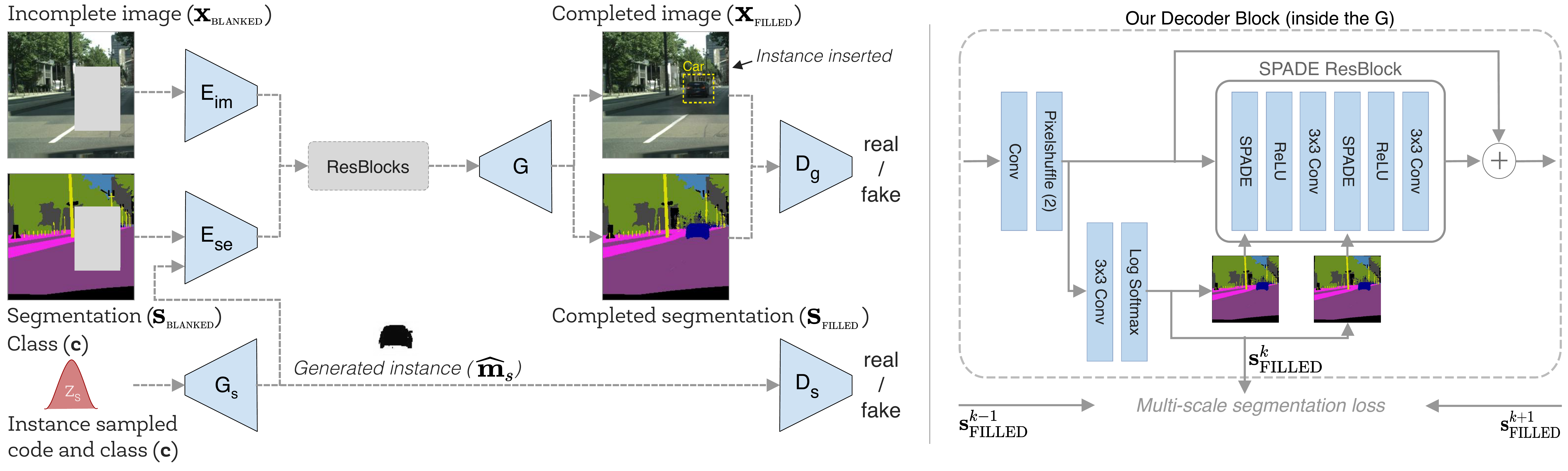}
   \caption{(left) Our model synthetizes a new image ($\bm{x}_{\text{\tiny FILLED}}$) and its segmentation ($\bm{s}_{\text{\tiny FILLED}}$) from an incomplete image ($\bm{x}_{\text{\tiny BLANKED}}$), its segmentation map ($\bm{s}_{\text{\tiny BLANKED}}$) and an optional instance, which might be sampled from a latent distribution ($\bm{z}\sim\mathcal{N}(0,\bm{I})$) or fed in input. The two discriminators encourage the generation of samples that resemble the real data distribution. (right) Our decoder block in the generator is based on SPADE~\cite{park2019semantic}. We propose a modification to use the predicted segmentation map instead of the ground truth segmentation and adopt a multi-scale segmentation loss to better learn it. }
   \label{fig:architecture}
\end{figure*}

However, synthesising new object instances in an existing complex scene is a challenging task as it involves the generation of a reliable structure and appearance, which has to fit harmoniously with the structure of the rest of the image. This problem has been much less studied in the literature.
Ouyang \textit{et al.}~\cite{ouyang2018pedestrian} propose a conditional GAN operating on the image manifold to insert plausible pedestrians into an urban scene. 
Hong \textit{et al.}~\cite{hong2018learning} instead suggest to focus on the semantic segmentation, inpaint missing regions and paste into them a segmentation mask provided by the user as input.
Lee \textit{et al.} \cite{lee2018context} propose a network that predicts the location and the shape of different objects (namely pedestrian and car) to insert them in the semantic segmentation space. Finally, Berlincioni \textit{et al.}~\cite{berlincioni2019road} focus on semantic segmentation removing cars and pedestrians from a road layout by feeding the binary masks in input to the model.
However, existing work allow users to either insert or delete objects from an existing scene. Interestingly, most of the literature focus just on the semantic segmentation space leaving to another network (e.g. pix2pix~\cite{isola2017image}) the generation of real pixels. 
This results in a fragmented approach that makes it difficult to edit an existing image. To the best of our knowledge, our approach is the first which permits to model multiple object instances' types and which allows, at the same time, to remove and reconstruct portions of images.

\vpara{From layout to images.} Noteworthy are the preliminary results in the related task of layout-to-image generation, which starts from a sparse object layout to synthetize plausible images. 
Pavllo \textit{et al.}~\cite{sun2020learning} created a layout-to-mask-to-image system starting from a bounding box layout to generate plausible segmentation maps and images. Zhao \emph{et al.}~\cite{zhao2019image} started from a scene description graph to generate possible images that correspond to the input graph. 
These works focus on synthetizing plausible images from an empty frame either using segmentation masks or bounding box layouts. However, in our setting we want to generate new objects and reconstruct images that have to fit and adapt well in the existing complex scene. Moreover, most of the layout-to-mask-to-image works test only few semantics and objects, while in our setting there are on average 17 different object categories per image.

\section{Approach}
In this paper, we aim at editing an image by removing and inserting object instances (e.g. car or pedestrian) in an existing urban context.
At inference time, users can thus remove existing objects by blanking-out a portion of the image and asking the network to insert an object instance to replace the removed one (see \Cref{fig:teaser}).

Inspired by recent literature of image inpainting, we propose a one-stage deep architecture that predicts the missing parts of the given images but also inserts new object instances. Our model allows the user to specify the exact position of the object or to let the network decide where to insert it inside a bigger area. 
We learn a latent space of object shapes from which sample the plausible objects to be inserted.
To help the network dealing with this challenging task, we guide the generation of new images through semantic segmentation.
\Cref{fig:architecture} shows an overview of the proposed network.


Formally, given an image $\bm{x}_{\text{\tiny BLANKED}} \in \mathbb{R}^{H\times W\times 3}$, its segmentation map $\bm{s}_{\text{\tiny GT}} \in\mathbb{R}^{H\times W\times C}$ ($C$ is the number of classes in the segmentation map), a latent code $\bm{z} \sim \mathcal{N}(0, \bm{I})$ and a one-hot encoding label $\bm{c}\in\mathbb{R}^D$, we want to synthetize a new image $\bm{x}_{\text{\tiny FILLED}}$. Instead of the missing area, this image has to contain an object instance of class $\bm{c}$ and to have the missing part reconstructed resembling the ground truth image $\bm{x}_{\text{\tiny GT}}\in \mathbb{R}^{H\times W\times 3}$, which is the original image without the blanked-out area.


\vpara{Training the network.} For a pair of images ($\bm{x}_{\text{\tiny GT}}$, $\bm{s}_{\text{\tiny GT}}$) in the dataset, we randomly generate a binary image mask $\bm{m} \in \{0,1\}^{H \times W \times 1}$ that defines the area of the original image to be blanked out. 
Whenever the mask $\bm{m}$ contains an object belonging to the modelled classes (e.g. car and pedestrian) we extract also the binary mask $\bm{m}_s \in \{0,1\}^{H \times W \times 1}$ that is $0$ everywhere except in the pixel locations of the object. We note that the mask $\bm{m}$ might be much bigger than the object mask $\bm{m}_s$.
Then, we use the image mask $\bm{m}$, to define $\bm{x}_{\text{\tiny BLANKED}} = \bm{x}_{\text{\tiny GT}} \odot \bm{m}$ and $\bm{s}_{\text{\tiny BLANKED}} = \bm{s}_{\text{\tiny GT}} \odot \bm{m}$. 

\vspace{0.05in}\noindent To ease the description of our method, we split it in two parts: object generation and inpainting, which are described as follows.

\subsection{Learning to generate instances}
We aim at inserting new objects in the semantic space.
Given an object $\bm{m}_{s}$, we want to learn a generative function starting from a latent code and a class generating the object $G_s(\bm{z}, \bm{c}) = \bm{m}_{s}$.
As inserting a new object is an ambiguous task (e.g. pedestrians might have different poses or shapes), we want to learn to generate multiple plausible objects and results from which the user can choose from.
Thus, we focus on learning a latent distribution of object shapes through an encoder network $E_s$, which will be then used to synthetize a new object shape in a VAE-fashion~\cite{kingma2013auto}. The VAE allows the generation of multiple objects by sampling multiple times from the learned latent distribution.

The shape and size of each object shape depends on its type (e.g. pedestrian) but also on its location in the scene. For example, pedestrians on the lower left part of the image are usually bigger than pedestrians on the upper part of the image, due to the different perspective.
Thus, we formulate the problem as learning an encoder $\bm{z} \sim E_s(\bm{m}_{s}, \bm{c}, \bm{l})$, where $\bm{l}$ is the location vector of the shape. 

Inspired by VAE literature~\cite{kingma2013auto}, we assume a low-dimensional latent space distributed as a Gaussian, from which we can sample $\bm{z} \sim \mathcal{N}(0, \bm{I})$ where $\bm{I}$ is the identity matrix.
Finally, we learn the encoder to learn the distribution with:
\begin{equation}
    \mathcal{L}_s^{VAE} = - \mathcal{D}_{KL}(E_s(\bm{m}_{s}, \bm{c}, \bm{l})\|\mathcal{N}(0,\bm{I}))
\end{equation}
where $\mathcal{D}_{KL}(p\|q) = -\int p(t)\log\frac{p(t)}{q(t)}dt$ is the Kullback-Leibler divergence.
This loss is ultimately expected to lead at learning the true posteriors of the latent distribution. 

Then, we encourage the generated mask $\bm{\hat{m}}_{s}$ to be as similar as possible to the original one after the sampling.
\begin{equation}
    \mathcal{L}_s^{rec} = \left \| \bm{m}_s - \bm{\hat{m}}_{s}\right \|_1
\end{equation}

We learn to generate realistic instances in an adversarial Least Square GAN~\cite{mao2017least}, i.e. considering a loss:
\begin{equation}
    \begin{aligned}
    \mathcal{L}_s^{adv}(G_s,D_s) = &\frac{1}{2}\mathbb{E}_{\bm{m}_s}[(D_s(\bm{m}_s))^2] + \frac{1}{2}\mathbb{E}_{\bm{\hat{m}}_{s}}[(D_s(\bm{\hat{m}}_{s}) - 1)^2]
    \end{aligned}
\end{equation}
where $D_s$ is the discriminator of object instances, based on DCGAN~\cite{radford2015unsupervised}.

We jointly train the generator $G_s$ and discriminator $D_s$
using the following objective function:
\begin{equation}
	\mathcal{L}^{adv}_s(G_s,D_s)= \lambda^{VAE}_s \mathcal{L}^{VAE}_s + \lambda^{rec}_s \mathcal{L}^{rec}_s  + \lambda^{adv}_s\mathcal{L}^{adv}_s
\end{equation}

\subsection{Learning to inpaint}
The goal of our model is not only to insert a new instance in the image, but also to modify the surrounding part by either altering the shape of existing objects, or removing them. 
Thus, we aim to complete the blanked image $\bm{x}_{\text{\tiny BLANKED}}$ and insert the object instance $\bm{\hat{m}}_{s}$ in the urban scene.
As the task of inserting a new object operates on the semantic space, we facilitate the network by using the semantic segmentation, which is often available either through large scale human annotated datasets~\cite{cordts2016cityscapes, varma2019idd, MVD2017} or semi/weakly/un-supervised approaches~\cite{wang2018high, li2018weakly, kalluri2019universal}.  

Given a blanked image $\bm{x}_{\text{\tiny BLANKED}}$, its corresponding segmentation $\bm{s}_{\text{\tiny BLANKED}}$, and a desired object class $\bm{c}$, the network has to output an image $\bm{x}_{\text{\tiny FILLED}}$ and its corresponding segmentation $\bm{s}_{\text{\tiny FILLED}}$, which have the blanked part reconstructed and with the object instance inserted.
Following \Cref{fig:architecture}, we feed the network with the three inputs and we sample $\bm{z}$ from the latent distribution. $E_{im}$ encodes the image $\bm{x}_{\text{\tiny BLANKED}}$, while $E_{se}$ encodes both $\bm{s}_{\text{\tiny BLANKED}}$ and the generated instance $\bm{\hat{m}}_{s}$. We note that, since $\bm{\hat{m}}_{s}$ is a binary image, the network has to understand the object class from its shape.
The intuition behind the two-streams encoder is that $E_{im}$ focuses on learning the style of the image, while $E_{se}$ focuses on the content and semantic by means of the semantic segmentation. 

The two encoders are then jointly fused in the feature level through a composition of several Residual Blocks \cite{he2016deep}. Then, in the decoder of the network, the image inpainting and semantic segmentation are progressively generated and updated across various decoder blocks with the segmentation map used to guide the inpainting of the masked image.

\vpara{Segmentation reconstruction and decoder block.} 
Motivated by the impressive results of the SPADE~\cite{park2019semantic} normalization on image generation from semantic segmentation, we propose a new decoder block to guide the generation process.
In the original formulation, SPADE exploits a human drawn semantic segmentation to synthetize photorealistic images.
Here, we instead guide the normalization with a predicted segmentation map, thus without relying on human drawn semantic segmentations.
This choice has two prominent advantages: i) it does not require the segmentation of the desired outcome (as we want to inpaint, we do not have it), and ii) it alleviates the mode collapse of the generated results that permits the nice consequence to have multiple diverse images.

\Cref{fig:architecture} (right) shows our decoder block. First, we get as input the features of the previous layer and upsample them using an upsampling sub-pixel convolution~\cite{shi2016real}.
These upsampled features are used to predict the segmentation map $\bm{s}^l_{\text{\tiny FILLED}}$ and forwarded to the SPADE Resblock~\cite{park2019semantic}. 
To remove any dependency of the batch we also replace SPADE Batch Normalization in favour of Instance Normalization~\cite{ulyanov2016instance}. 

To encourage consistency of real pixels and segmentation masks, but also of segmentation masks across the decoder layers, we use a multi-scale segmentation loss between the predicted segmentation mask and the ground truth
\begin{equation}
\begin{aligned}
    \mathcal{L}_c^{se} = - \sum_{k}^C \sum_{i=0}^{m} \bm{s}^k_{\text{\tiny GT}} \log(\textit{upscale}(\bm{s}^k_i))
    \end{aligned}
\end{equation}
where $C$ is the number of considered semantics (e.g. road, car), and \textit{upscale} interpolates $\bm{s}_i$ to match the size of $\bm{s}_{\text{\tiny GT}}$ and compute $\mathcal{L}_c^{se}$ on highly detailed segmentation maps.

\vpara{Image reconstruction.} We encourage the network at reconstructing the image with different state of the art losses:
\begin{itemize}[leftmargin=*]
	\item \emph{Pixels reconstruction.} Given an image sampled from the data distribution, we should be able to reconstruct it after encoding and decoding. This can be obtained using:
	\begin{equation}
    \begin{aligned}
        \mathcal{L}_{c}^{rec} = &\left \| \bm{x}_{\text{\tiny FILLED}} - \bm{x}_{\text{\tiny GT}}\right \|_1
        \end{aligned}
    \end{equation}
    The $\mathcal{L}_1$ loss encourages the generation of sharper images than the $\mathcal{L}_2$~\cite{isola2017image}. The loss is normalised by the mask size.
    
    \item \emph{Feature-Matching Loss.} Inspired by~\cite{wang2018high}, the feature-matching loss compares the activation maps of the layers of the discriminator. This forces the generator to produce images whose representations are similar to the ground truth in the discriminator space. It is defined as:
    \begin{equation}
        \begin{aligned}
        \mathcal{L}_c^{FM} = \mathbb{E} \bigg[\sum_{k=1}^K \sum_{i= 1}^{L} \frac{\| D_k^{(i)}(\bm{x}_{\text{\tiny GT}}^{k},\bm{s}_{\text{\tiny GT}}^{k}) - D_k^{(i)}(\bm{x}_{\text{\tiny FILLED}}^{k}, \bm{s}_{\text{\tiny GT}}^{k})   \|_1}{K}\bigg]
        \end{aligned}
    \end{equation}
    where $K$ is the number of scales of the discriminator, $L$ is the last layer of the discriminator and $D_k^{(i)}$ is the activation map of the $i_{th}$ layer of the discriminator.
    
    \item \emph{Perceptual and Style Loss.} First introduced in \cite{gatys2016image, johnson2016perceptual}, $\mathcal{L}_{c}^{perc}$ is used to penalize results that are not perceptually similar to the source image, in this case the ground truth, and it is particularly important in complex scenes, where multiple details and objects are present. The perceptual distance is measured by the distance between the activation maps of the two images using a pretrained network. Formally:
    \begin{equation}
        \mathcal{L}_{c}^{perc} = \mathbb{E}\bigg[\sum_{i=1}^{4} \frac{1}{N_i}\left \|\phi_l (\bm{x}_{\text{\tiny FILLED}}) - \phi_l(\bm{x}_{\text{\tiny GT}})  \right \|_1 \bigg]
    \end{equation}
    where $\phi_i$ corresponds to the activation map of the $i$-th layer of a ImageNet pre-trained VGG-19 network~\cite{simonyan2014very}. $N_i = H_jW_jC_j$ is a normalization factor that takes into account the number of elements of VGG-19.
    The activation maps of the VGG-19 network are also used to compute the style loss, $\mathcal{L}_{c}^{style}$, which measures the differences between covariances of the activation maps and penalizes the style shifting of the two images. Formally:
    \begin{equation}
        \mathcal{L}_{c}^{sytle} = \mathbb{E}_j \bigg[ \left \| (G_j^{\phi}(\bm{x}_{\text{\tiny FILLED}}) - G_j^{\phi}(\bm{x}_{\text{\tiny GT}}))\right \|_1 \bigg]
    \end{equation}
    where $G_j^{\phi}$ is computed as $G_j^{\phi} = \psi\psi^{T}\frac{1}{N_i}$, $\psi$ is $\phi_j$ reshaped into a $C_j \times H_jW_j$ matrix. Similarly to $\mathcal{L}_{c}^{perc}$, the style loss encourage high quality results in complex scenes.
\end{itemize}

We learn the generation through an adversarial with a Least Square loss~\cite{mao2017least}.
We construct the adversarial game with a multiscale discriminator PatchGAN~\cite{wang2018high, isola2017image}
\begin{equation}
    \begin{aligned}
        \mathcal{L}_c^{adv}(G_s,D_s) = \sum^K_{k=1} &\frac{1}{2} \mathbb{E}_{\bm{x}_{\text{\tiny GT}}}[( D_g^k(\bm{x}^k_{\text{\tiny GT}}, \bm{s}^k_{\text{\tiny GT}}))^2] + \\ & \frac{1}{2} \mathbb{E}_{\bm{x}_{\text{\tiny FILLED}}}[( D_g^k(\bm{x}^k_{\text{\tiny FILLED}}, \bm{s}^k_{\text{\tiny GT}}) -1 )^2] 
    \end{aligned}
\end{equation}
where $K$ is the number of scales and $k$ refers to the $k^{th}$ scale of the image and segmentation map. Each layer $k$ of the discriminator $D_g$ takes as input the generated image $\bm{x}^k_{\text{\tiny FILLED}}$ and the ground truth segmentation map $\bm{s}^k_{\text{\tiny GT}}$. The ground truth segmentation map $\bm{s}^k_{\text{\tiny GT}}$ is here useful to verify the coherence between the generated image and the semantic segmentation.

The entire inpaint network is trained with the following losses:
\begin{equation}
    \mathcal{L}_{D_c} = \mathcal{L}_c^{adv}(G_c,D_c)
\end{equation}
\begin{equation}
    \begin{aligned}
        \mathcal{L}_{G_c} = &\mathcal{L}_c^{adv}(G_c,D_c) +  \lambda_c^{rec}\mathcal{L}_{c}^{rec}(G_c) + 
    \lambda_c^{perc}\mathcal{L}_{c}^{perc}(G_c) + \\& \lambda_c^{style}\mathcal{L}_{c}^{style}(G_c) + \lambda_c^{FM}\mathcal{L}_{c}^{FM}(G_c) + \lambda_c^{cross}\mathcal{L}_{c}^{cross}(G_c)
    \end{aligned}
\end{equation}
where $\lambda_{rec}$, $\lambda_{perc}$, $\lambda_{style}$, $\lambda_{FM}$ and $\lambda_{cross}$ are hyper-parameters for the weights of the corresponding loss terms. The value of most of these parameters come from the literature. We refer to Supplementary for the details.


\section{Experiments}

We conduct a quantitative evaluation on two widely recognized datasets for urban benchmarks, namely Cityscapes~\cite{cordts2016cityscapes} and the Indian Driving~\cite{varma2019idd}. The former is focused on European contexts while the latter exhibits a higher diversity of pedestrians and vehicles, but also ambiguous road boundaries and conditions. 
For both the datasets we first resize the images to height 256, then we random crop them to 256 $\times$ 256. Inspired by literature~\cite{cordts2016cityscapes}, we aggregate the 35 segmentation map categories into 17 groups for Cityscapes, and in Indian Driving we follow a similar approach from 40 categories to 21 groups.
We refer to the Supplementary for additional details.

\subsection{Baselines}
As baselines we select three state of the art models for image manipulation, namely Hong \emph{et al}~\cite{hong2018learning}, and inpainting, i.e. SPG-Net~\cite{song2018spg} and RN~\cite{yu2019region}. Hong \emph{et al}~\cite{hong2018learning} is a two-stage manipulation model that uses the segmentation mask and insert or remove objects from an image. The user is required to specify the exact bounding box where to insert the object. SPG-Net is a two-stage inpainting network that leverages the semantic segmentation as input, while RN is a one stage inpainting network learning a normalization layer to achieve consistent performance improvements over the previous works.
We used the source code released by RN authors, while we implemented from scratch SPG-Net following the description of the original paper as the code was not available.
We release the code of our model and the implementation of SPG-Net.
 
\subsection{Experimental settings}

We design the following two testing setups for our network.
 
\vpara{Restore.} We test the models for image inpainting, where a portion of the image is blanked out and the network has to generate a plausible portion of the image.

\vpara{Place.} We test the models for the task of object insertion and image inpainting together. Thus, we blank out a portion of the image. The networks have to reconstruct the image and the desired object have to be present in the reconstruction. As SPG-Net and RN are not able to insert objects, we modify them to use an instance sampled from the latent space trained with the same VAE we use in our proposal.
We call these networks SPG-Net* and RN*.
Since Hong \emph{et al} ~\cite{hong2018learning} is only able to place objects as big as the bounding box, we first restore the missing part, then we ask the network to place a new object in the exact position where the ground truth object is. We note that this setting might favour Hong \emph{et al}~\cite{hong2018learning} model.

\vspace{0.05in}\noindent In order to conduct a fair comparison with ~\cite{song2018spg}, for the \emph{restore} task we use rectangular masks. Thus, for each image, we create a single rectangular mask at random locations. 
Each mask can have a size that goes from 32x32 up to 128x128.  
For the \emph{place} task, we maintain the same size of the mask of the \emph{restore} one but we change the position. In particular, we firstly extract a valid instance from the instance-wise annotation of each image. We pre-process the dataset extracting information for each object discarding instances that are too small or occluded from other objects. Then, we randomly choose one from the list of valid objects and we generate the mask based on the position of the instance. Whenever an image does not contain a valid object, we generate a new mask at random position.
We release the generated mask to ease the comparison of future research with our model.

\begin{table}[!ht]
    \centering
    \caption{Quantitative results for our model and the baselines. We evaluate the models through image quality (PSNR and FID) and accuracy of instance insertion (F1).}
    \label{tab:results}
    \begin{tabular}{@{}clc rrr r rrr@{}}
    \midrule
    & \multirow{2}{*}{\textbf{Model}} &
    \multicolumn{3}{c}{\textbf{Cityscapes}} &  \multicolumn{3}{c}{\textbf{Indian Driving}} \\
    \cmidrule(l{2pt}){3-5} \cmidrule(l{2pt}){6-8}
    && PSNR$\uparrow$ & FID$\downarrow$ & F1$\uparrow$ & PSNR$\uparrow$ & FID$\downarrow$ & F1$\uparrow$\\
    \midrule
    \multirow{4}{*}{\rotatebox{90}{Restore}} & Hong \emph{et al.}~\cite{hong2018learning} & 31.07 & 7.26 & 0.00 & 30.31 & 6.34 & 0.00\\
    & SPG-Net~\cite{song2018spg} & 31.36 & 7.97 & 0.00 & 29.95 & 6.39 & 0.02\\
    & RN~\cite{yu2019region}  & 32.16 & 9.64 & 0.00 & 29.83 & 11.14 & 0.02\\
    & Our proposal & \textbf{32.95} & \textbf{5.08} & \textbf{0.06} & \textbf{30.97} & \textbf{5.45} & \textbf{0.05}\\
    \midrule
    \multirow{4}{*}{\rotatebox[origin=l]{90}{Place}} & Hong \emph{et al.}~\cite{hong2018learning} & 31.08 & 7.26 & 0.10 & 30.32 & 6.32 & 0.91\\
    & SPG-Net* & 31.37 & 7.96 & 0.60 & 29.94 & 6.38 & 0.87\\
    & RN*  &  31.74 & 9.79 & 0.54 & 29.62 & 10.76 & 0.71\\
    & Our proposal &  \textbf{32.96} & \textbf{5.05}  & \textbf{0.91} & \textbf{30.98} & \textbf{5.43} & \textbf{0.97} \\
    \bottomrule      
    \end{tabular}
\end{table}

\begin{figure*}[ht]
    \setlength{\tabcolsep}{1pt}
	\renewcommand{\arraystretch}{0.8}
    \newcommand{\sizea}{0.131\linewidth}
    \footnotesize
	\centering
	\begin{tabular}{ccc ccccc c}
	& & Input image & Hong \textit{et al.}~\cite{hong2018learning} & SPG-Net* & RN* & Our & Our (segmentation) & Ground truth \\
	   \multirow{2}{*}{\rotatebox[origin=c]{90}{\small \textbf{Cityscapes}}} & \rotatebox[origin=c]{90}{\small Car} & 
	   \raisebox{-0.5\height}{\includegraphics[width=\sizea]{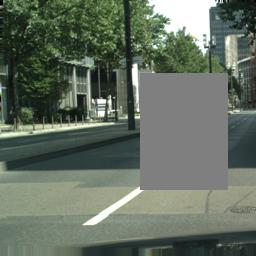}} &
	    \raisebox{-0.5\height}{\includegraphics[width=\sizea]{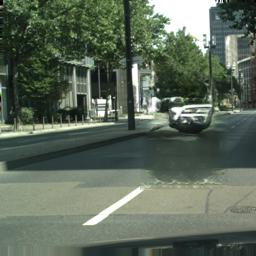}} &  
       \raisebox{-0.5\height}{\includegraphics[width=\sizea]{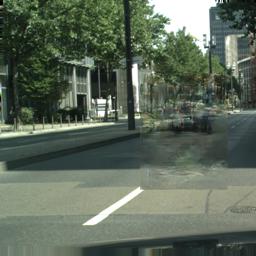}} &
	   \raisebox{-0.5\height}{\includegraphics[width=\sizea]{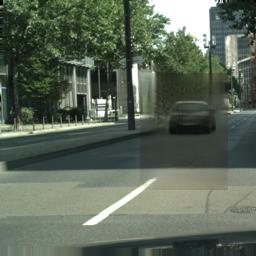}} &
	   \raisebox{-0.5\height}{\includegraphics[width=\sizea]{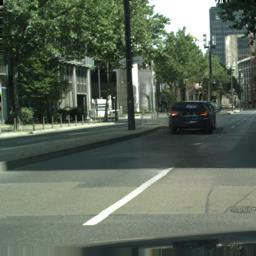}} &
	   \raisebox{-0.5\height}{\includegraphics[width=\sizea]{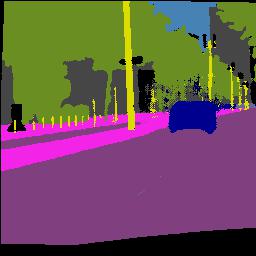}} &
	   \raisebox{-0.5\height}{\includegraphics[width=\sizea]{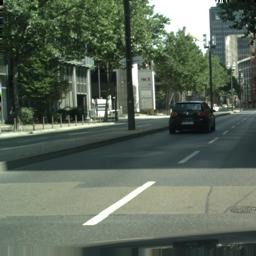}} \\ 
	   & \rotatebox[origin=c]{90}{\small Pedestrian} & \raisebox{-0.5\height}{\includegraphics[width=\sizea]{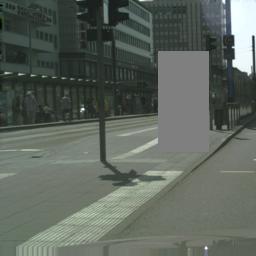}} &
	   \raisebox{-0.5\height}{\includegraphics[width=\sizea]{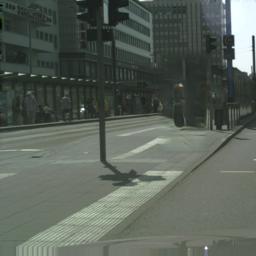}}
	   &
       \raisebox{-0.5\height}{\includegraphics[width=\sizea]{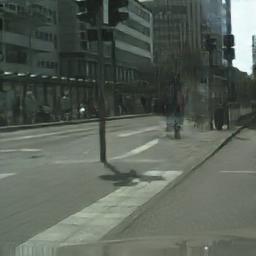}} &
	   \raisebox{-0.5\height}{\includegraphics[width=\sizea]{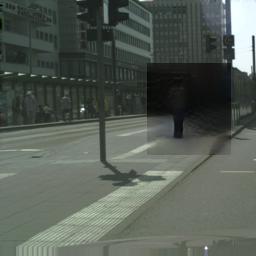}} &
	   \raisebox{-0.5\height}{\includegraphics[width=\sizea]{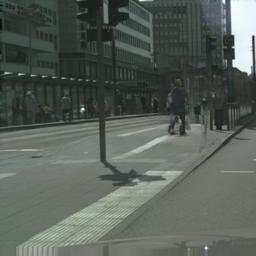}} &
	   \raisebox{-0.5\height}{\includegraphics[width=\sizea]{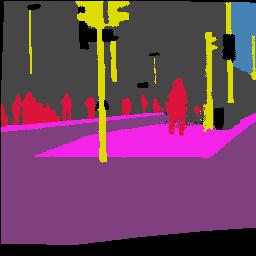}} &
	   \raisebox{-0.5\height}{\includegraphics[width=\sizea]{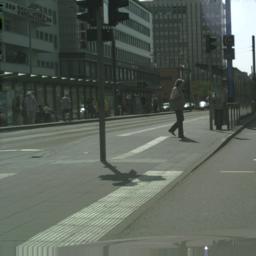}}\\ 
	   
	   \multirow{2}{*}{\rotatebox[origin=c]{90}{\enspace \small \textbf{Indian Driving}}} & \rotatebox[origin=c]{90}{\small Car} & \raisebox{-0.5\height}{\includegraphics[width=\sizea]{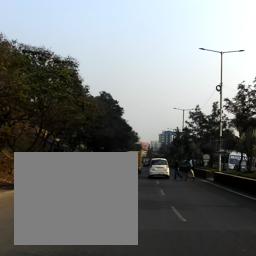}} &
	   \raisebox{-0.5\height}{\includegraphics[width=\sizea]{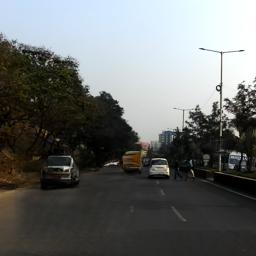}} &
       \raisebox{-0.5\height}{\includegraphics[width=\sizea]{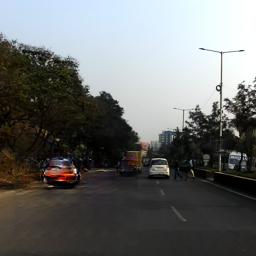}} &
	   \raisebox{-0.5\height}{\includegraphics[width=\sizea]{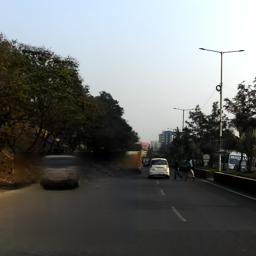}} &
	   \raisebox{-0.5\height}{\includegraphics[width=\sizea]{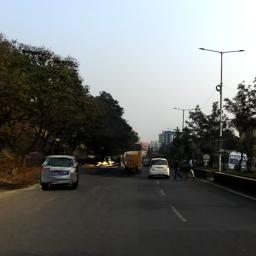}} &
	   \raisebox{-0.5\height}{\includegraphics[width=\sizea]{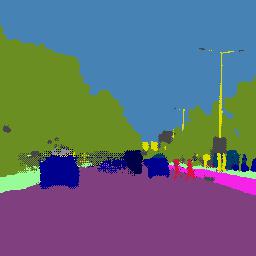}} &
	   \raisebox{-0.5\height}{\includegraphics[width=\sizea]{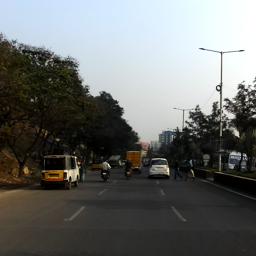}}\\ 
	   & \rotatebox[origin=c]{90}{\small Pedestrian} & \raisebox{-0.5\height}{\includegraphics[width=\sizea]{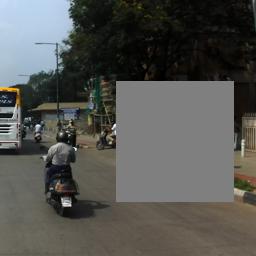}} & 
	   \raisebox{-0.5\height}{\includegraphics[width=\sizea]{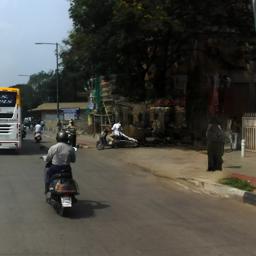}} &
       \raisebox{-0.5\height}{\includegraphics[width=\sizea]{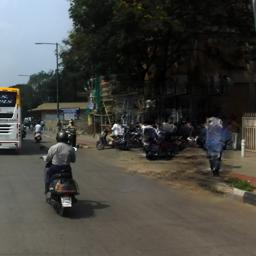}} &
	   \raisebox{-0.5\height}{\includegraphics[width=\sizea]{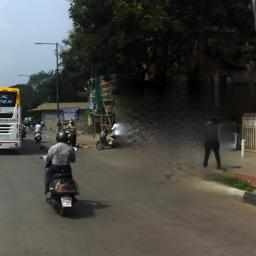}} &
	   \raisebox{-0.5\height}{\includegraphics[width=\sizea]{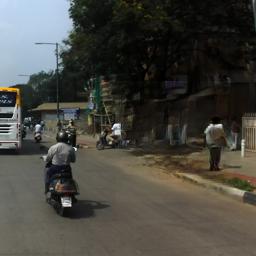}} &
	   \raisebox{-0.5\height}{\includegraphics[width=\sizea]{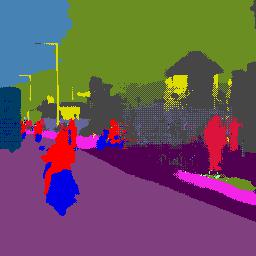}} &
	   \raisebox{-0.5\height}{\includegraphics[width=\sizea]{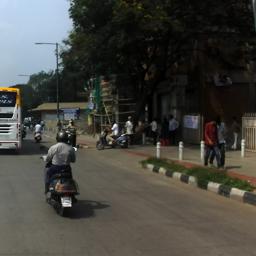}}\\ 
	\end{tabular}
	\caption{Qualitative evaluation on the task of object insertion and inpainting. The first two rows show results on the Cityscape dataset while the last two rows show the Indian Driving results. We show the results on the same multi-domain model conditioned on two types of object insertion: cars and pedestrians.}
	\label{Fig:objectInsertion}
\end{figure*}

\subsection{Evaluation}
We evaluate generated results through image quality and instance accuracy. We measure image quality through the Peak Signal-to-Noise Ratio (PSNR) and the Fr\'echet Inception Distance (FID)~\cite{heusel2017gans}. The FID is the Wasserstein-2 distance between the generated and real image feature representations extracted from a pre-trained Inception-V3 model~\cite{szegedy2016rethinking}. Higher PSNR and Lower FID values indicate higher image quality.

To measure the presence of inserted objects we use the F1 metric through the use of a pre-trained YOLOv3~\cite{redmon2018yolov3} network, which detects whether the desired instances are inserted. Higher scores indicate that the network is inserting synthetic objects that resemble the real ones.
For the \emph{restore} experimental setting, we compute the score by detecting whether at least one of the modelled instances (i.e. cars and pedestrians) is inserted.

\vpara{Perceptual user study.} We also run a perceptual study asking 19 users to conduct a user study on both datasets. Each user evaluates 38 random images, 19 for each experimental setting. For each image we show the original masked image and the generated output of all the networks. The users are asked to select the best output judging the realism and quality of reconstructed pixels.     

\section{Results}

\vpara{Quantitative results.}
We evaluate our proposal with the state of the art solutions in two settings: \emph{restore} and \emph{place}, which test the ability to inpaint missing portions of images and manipulate an image (insert, remove, reconstruct), respectively.
\Cref{tab:results} shows that our approach outperforms all the compared models in the two settings. 
In particular, we observe that our model significantly improves the quality of generated images, measured through FID and PSNR. 
In the \emph{restore} setting, we observe that, as expected, all inpaiting models collapse to object removal, rarely proposing the insertion of a new object in an existing scene.
Thus, we condition existing models to insert new objects (\emph{place} setting). Without significant losses on the image quality, we observe that both SPG-Net* and RN* learn to insert object instances. However, our proposal greatly improves both the image quality and the accuracy of object insertion, which increases on average by 18\% and 310\% in Indian Driving and Cityscapes, respectively.
Overall, we also show that merely changing existing models (RN and SPG-Net) does not achieve state of the art performance. 

\vpara{Qualitative results.}
We begin by commenting on the results of the most challenging \emph{place} experimental setting. 
\Cref{Fig:objectInsertion} shows four different images from which we want to reconstruct the missing part but also insert a user-defined object (e.g. car or pedestrian).
We can see that our method correctly inserts the required object, while others struggle at doing it, especially in Cityscapes.
While RN is often able to generate object shapes that are correctly detected by YOLOv3~\cite{redmon2018yolov3}, a visual inspection of the second and fourth row of \Cref{Fig:objectInsertion} highlights that the object is poorly colorized. Moreover, all the baselines fail at adapting the reconstructed pixel to the scene.
In particular, SPG-Net and Hong \emph{et al.} suffer from significant blurriness and artifacts (see \Cref{Fig:objectInsertion} first rows), while RN unreliably reconstructs the edges and result in blurry reconstructions.
On the contrary, our proposal results in sharper images and distinguishable inserted objects. 

\begin{figure}[ht]
    \setlength{\tabcolsep}{1pt}
	\renewcommand{\arraystretch}{0.8}
    \newcommand{\sizea}{0.19\linewidth}
    \footnotesize
	\centering
	\begin{tabular}{ccccccc c}
	Input image & Hong \textit{et al.}~\cite{hong2018learning} & SPG-Net & RN & Our \\
	   \includegraphics[width=\sizea]{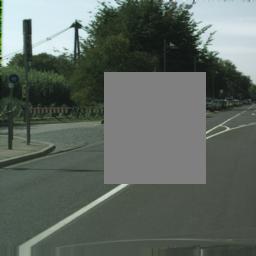} & \includegraphics[width=\sizea]{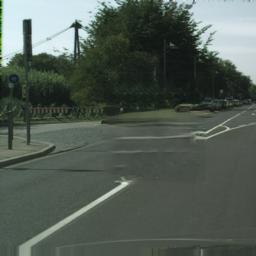} &
       \includegraphics[width=\sizea]{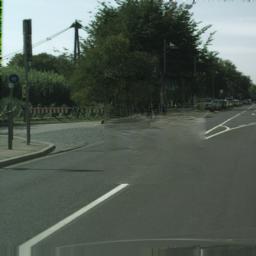} &
	   \includegraphics[width=\sizea]{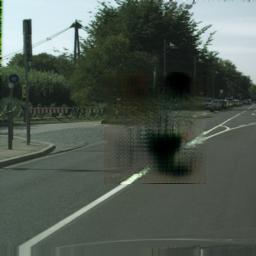} &
	   \includegraphics[width=\sizea]{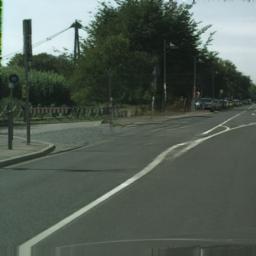} \\
	   \includegraphics[width=\sizea]{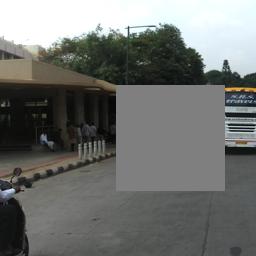} &
	   \includegraphics[width=\sizea]{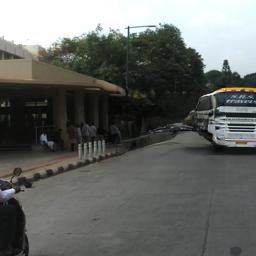} &
       \includegraphics[width=\sizea]{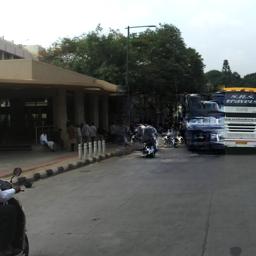} &
	   \includegraphics[width=\sizea]{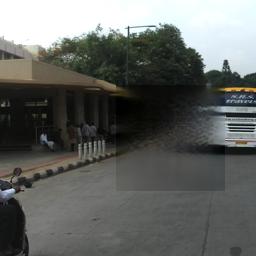} &
	   \includegraphics[width=\sizea]{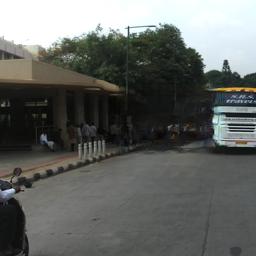} \\
	\end{tabular}
	\caption{Qualitative evaluation on the task of inpainting without object insertion. The first row shows results on the Cityscape dataset, while the last row shows the Indian Driving results. Zoom in for better details.}
	\label{Fig:Inpaiting}
\end{figure}

We now discuss the \emph{restore} results where the model has only to reconstruct missing parts of the image.
\Cref{Fig:Inpaiting} shows that our model generates sharper images with pixels that are well adapted in the original scene.
In particular, we observe that our model can reconstruct even the horizontal road marking quite well (see the first row \Cref{Fig:Inpaiting}).
State of the art models, instead, seem to struggle at the reconstruction. These poor results might be a consequence of two main reasons.
First, RN focuses on free-form missing parts, while we might generate big areas where their normalization might fail.
Second, the reason behind the ``checkerboard'' artifacts of SPG-Net might lie on the Deconvolution applied in the decoder in order to upsample the features. Indeed, Deconvolution has been shown to produce results that present various artifacts ~\cite{odena2016deconvolution} and that can be reduced using style loss~\cite{sajjadi2017enhancenet}.

Thus, we propose a new decoder block, based on SPADE, that jointly predicts the segmentation and real pixels. By using the predicted segmentation as input to the SPADE normalization, we guide the decoder to reconstruct semantics that are consistent with the real pixels. 
As a result, edges and contours are visually pleasing and better reconstructed than the state of the art models.
It is worth noting that our decoder can also be applied easily to other existing models.

Our holistic framework can be applied in a wide range of manipulations of complex scenes. Thus, we also test our model performance in inserting and removing an user specified input shape (see \Cref{fig:teaser} for the manipulation types). 
Qualitative and quantitative results for these two tasks show that our model significantly outperforms the baselines (see Supplementary Fig. S1 and S2). 
Surprisingly, we qualitatively observe that RN generates low quality results and visible boundaries even in the free-form task in which they focus, highlighting the challenge of complex scene manipulation.

\begin{table}[!h]
    \centering
    \caption{Ablation study of the architecture and input quality.}
    \label{tab:ablation}
    \begin{tabular}{@{}l rrrr@{}}
	\toprule
		\textbf{Model} & \textbf{PSNR$\uparrow$} & \textbf{FID$\downarrow$}\\
		\midrule
		Our proposal (A) &  \textbf{32.96} & \textbf{5.05}\\
	    (A) w/o $\mathcal{L}_{\textrm{style}}$ & 32.68 & 5.38\\
		(A) w/o $\mathcal{L}_{\textrm{FM}}$ & 32.66 & 5.30\\
		(A) w/o $E_{se}$ and $\bm{s}_{\text{\tiny BLANKED}}$ & 32.35 & 5.42\\
		(A) w/o SPADE & 32.57 & 5.56\\ 
		\midrule
		(A) using DeepLabv3~\cite{chen2017rethinking} segmentation & 32.85 & 5.11\\
    \bottomrule      
    \end{tabular}
    \vspace{-4mm}
\end{table}

\vpara{Perceptual user study.} The user study shows that our generated images have been selected 85\% of the time. Specifically, our results are perceptually better 90\% and 80\% of the time in the \textit{restore} task and \textit{place} tasks respectively. 

\vpara{Ablation.}
Our model introduces various novel components and different state of the art losses whose contribution has to be validated. 
We performed an ablation study on Cityscapes, which is the standard testbed for inpainting in complex scenes.

We begin by testing the contribution of the style $\mathcal{L}_{\textrm{style}}$ and feature matching $\mathcal{L}_{\textrm{FM}}$ losses. \Cref{tab:ablation} shows that removing one of the two losses results in a small performance drop. 
However, qualitative results show they reduce the ``checkerboard'' artifacts in the image, which is confirmed by recent literature~\cite{sajjadi2017enhancenet} (see the Supplementary materials).

Similar to SPG-Net, our network uses a segmentation mask in input, which is supposed to guide the network at learning a better feature representation. However, it is arguably a costly choice as it requires human annotations. Moreover, our decoder block predicts the segmentation and uses the SPADE normalization, which might be sufficient to have high-quality results.
To test these two hypotheses, we first train a network without the segmentation encoder $E_{se}$ and the segmentation map $\bm{s}_{\text{\tiny BLANKED}}$. From \Cref{tab:ablation}, we can observe a significant drop in performance, but our network still performs better than state of the art without these two components. Thus, this setting might be considered in limited-resource scenarios.
Then, we test the use of predicted segmentation maps instead of the ground truth maps. So, we use segmentation images generated by a pre-trained segmentation predictor DeepLabv3~\cite{chen2017rethinking}. We observe no significant performance degradation.

Finally, we observe that by removing the entire SPADE block we see a significant negative effect on image quality. However, these results are better than the state of the art models, which show that the contribution of our model is not only a consequence of SPADE. We refer to the Supplementary for additional ablations for components of our SPADE block.

\section{Conclusions}
In this paper, we proposed a novel framework that unifies and improve previous methods for image inpainting and object insertion.
At inference time, users interact with our model by feeding as input an image and its segmentation that have blanked area, and optionally an object (label or shape) to be inserted. The model encodes both the two input images, sample from a latent distribution whenever a class label is fed, and reconstruct the images. If users desire to insert the object, the reconstructed images contains it as expected.

To encourage high quality reconstructions, we use semantic segmentation maps both in input and in the decoder. Specifically, we proposed a new decoder block that is based on SPADE and mixes a semantic prediction task with the normalization to generate adequate images. We believe that this decoder block might be applied to a wide range of tasks that exploit the semantic segmentation to generate new images.

To the best of our knowledge, we are the first at learning to jointly remove, inpaint and insert objects of multiple types in a single one-stage model. We hope that our work will stimulate future research on the challenging manipulation of complex scenes, having multiple cluttered objects and semantic classes.

\bibliographystyle{IEEEtran}
\bibliography{acmart}

\clearpage
\onecolumn
\appendix
\section{Implementation Details}
\label{suppl:implementation_details}
The encoder of our image completion architecture follows the two-stream convolutional encoder of \cite{hong2018learning}. Specifically, both encoders downsample the input four times, using Instance normalization \cite{ulyanov2017improved} and ELU \cite{clevert2015fast} as activation function. 
The encoder is followed by nine residual blocks. With respect to the original proposed residual block, we have decided to use dilated convolution with increasing dilation factor as proposed in \cite{yu2017dilated}, Instance normalization \cite{ulyanov2017improved}, ELU activation \cite{clevert2015fast}. Moreover we have decided to remove the activation function at the end of each residual block after an evaluation phase.
Out proposed model has been implemented using PyTorch. The batch has been set to four due to GPU limitation. The model has been trained for 600 epochs for the Cityscapes dataset and for 200 epochs for the Indian Driving Dataset using Adam~\cite{kingma2014adam} as optimizer with a learning rate of 0.0002. For our experiments we choose $\lambda_{recon} = \lambda_{perc} = \lambda_{FM} = \lambda_{cross} $= 10,  $\lambda_{style}$ = 250, $\lambda_{VAE} $= 5 and $\lambda_{inst\_recon} $= 20. Spectral normalization (SN) ~\cite{miyato2018spectral} was used in in the discriminator. 
Although recent works~\cite{zhang2018self} have shown that SN can be used also for the generator, it further increases the training time with only minor improvements. For this reason, we have decided to not use it. The details of the architecture are shown in \Cref{tab:architecture} and \Cref{tab:architecture_1}. We release the source code of our model and the SPG-Net implementation at \texttt{https://github.com/PierfrancescoArdino/SGINet}.

\begin{table*}[!ht]
	\centering
	\begin{tabular}{@{}lll@{}}
		\toprule
		\textbf{Part} & \textbf{Input} $\rightarrow$ \textbf{Output Shape} & \textbf{Layer Information} \\ \midrule
		\multirow{8}{*}{$E_s$} & (4104,) $\rightarrow$ (8128,) & FC-(8128), LeakyReLU \\
		& (, 8128) $\rightarrow$ (32, 32, 8) & RESHAPE \\
		& (32, 32, 8) $\rightarrow$ (32, 32, 32) & CONV-(N32, K3x3, S1, P1), IN, LeakyReLU \\ 
		& (32, 32, 32) $\rightarrow$ (16, 16, 64) & CONV-(N64, K4x4, S2, P1), IN, LeakyReLU \\ 
		& (16, 16, 64) $\rightarrow$ (8, 8, 128) & CONV-(N128, K4x4, S2, P1), IN, LeakyReLU \\ 
		& (8, 8, 128) $\rightarrow$ (4, 4, 256) & CONV-(N256, K4x4, S2, P1), IN, LeakyReLU \\ 
		\cmidrule{2-3}
		& (4,4, 256) $\rightarrow$ ($Z$,) & CONV-($Z$, K4x4, S1, P0)\\
		& (4,4 256) $\rightarrow$ ($Z$,) &  CONV-($Z$, K4x4, S1, P0)\\
		\midrule
		\multirow{5}{*}{$G_s$} & ($Z$ + $N_{CLASS}$ + $\theta$) $\rightarrow$ ($\frac{h}{16}$, $\frac{h}{16}$, 256) & DECONV-(N256, K4x4, S1, P0), IN, ReLU \\
		& ($\frac{h}{16}$, $\frac{h}{16}$, 256) $\rightarrow$ ($\frac{h}{8}$, $\frac{h}{8}$, 128) & DECONV-(N128, K4x4, S2, P1), IN, ReLU \\
		& ($\frac{h}{8}$, $\frac{h}{8}$, 128) $\rightarrow$ ($\frac{h}{4}$, $\frac{h}{4}$, 64) & DECONV-(N64, K4x4, S2, P1), IN, ReLU \\ 
		& ($\frac{h}{4}$, $\frac{h}{4}$, 64) $\rightarrow$ ($\frac{h}{2}$, $\frac{h}{2}$, 32) & DECONV-(N32, K4x4, S2, P1), IN, ReLU \\
		& ($\frac{h}{2}$, $\frac{w}{2}$, 32) $\rightarrow$ ($h$, $w$, 1) & CONV-(N32, K4x4, S2, P1), SIGMOID \\
		\midrule
		\multirow{5}{*}{$D_s$} & ($h$,$w$,1) $\rightarrow$ ($\frac{h}{2}$,$\frac{w}{2}$,64) & CONV-(N64, K4x4, S2, P1), LeakyReLU \\ 
		& ($\frac{h}{2}$,$\frac{w}{2}$,64) $\rightarrow$ ($\frac{h}{4}$,$\frac{w}{4}$,128) & CONV-(N128, K4x4, S2, P1), IN, LeakyReLU \\ 
		& ($\frac{h}{4}$,$\frac{w}{4}$,128) $\rightarrow$ ($\frac{h}{8}$,$\frac{w}{8}$,256) & CONV-(N256, K4x4, S2, P1), IN, LeakyReLU \\ 
		& ($\frac{h}{8}$,$\frac{w}{8}$,256) $\rightarrow$ ($\frac{h}{16}$,$\frac{w}{16}$,512) & CONV-(N512, K4x4, S2, P1), IN, LeakyReLU \\ 
		& ($\frac{h}{16}$,$\frac{w}{16}$,256) $\rightarrow$ (1,1,1) & CONV-(N1, K4x4, S1, P0) \\
		\bottomrule
	\end{tabular}
	\vspace{1mm}
	\caption{Network architecture. We use the following notation: $Z$: the dimension of attribute vetor, $N_{CLASS}$: the number of the class instance, $\theta$: the number of location parameters of the affine transformation, K: kernel size, S: stride size, P: padding size, CONV: a convolutional layer, DECONV: a deconvolutional layer, FC: fully connected layer, IN: Instance Normalization }
	\label{tab:architecture}
\end{table*}

\begin{table*}[!ht]
	\centering
	\begin{tabular}{@{}lll@{}}
		\toprule
		\textbf{Part} & \textbf{Input} $\rightarrow$ \textbf{Output Shape} & \textbf{Layer Information} \\ \midrule
		\multirow{5}{*}{$E_{im}$} & (${h}$, ${w}$, 3 + $\bm{m}$) $\rightarrow$         (${h}$, ${w}$, 32) & CONV-(N32, K5x5, S1, P2), IN, ELU \\ 
		& (${h}$, ${w}$, 32) $\rightarrow$ ($\frac{h}{2}$, $\frac{w}{2}$, 64) & CONV-(N64, K5x5, S1, P2), IN, ELU \\
		& ($\frac{h}{2}$, $\frac{w}{2}$, 64) $\rightarrow$ ($\frac{h}{4}$, $\frac{w}{4}$, 128) & CONV-(N128, K4x4, S2, P1), IN, ELU \\
		& ($\frac{h}{4}$, $\frac{w}{4}$, 128) $\rightarrow$ ($\frac{h}{8}$, $\frac{w}{8}$, 256) & CONV-(N256, K4x4, S2, P1), IN, ELU \\ 
		& ($\frac{h}{8}$, $\frac{w}{8}$, 256) $\rightarrow$ ($\frac{h}{16}$, $\frac{w}{16}$, 512) & CONV-(N512, K4x4, S2, P1), IN, ELU \\
		\midrule
		\multirow{5}{*}{$E_{se}$} & (${h}$, ${w}$, $\mathcal{C} + \bm{m} + \mathbf{\hat{m}}_{s}$) $\rightarrow$ (${h}$, ${w}$, 32) & CONV-(N32, K5x5, S1, P2), IN, ELU \\ 
		& (${h}$, ${w}$, 32) $\rightarrow$ ($\frac{h}{2}$, $\frac{w}{2}$, 64) & CONV-(N64, K5x5, S1, P2), IN, ELU \\
		& ($\frac{h}{2}$, $\frac{w}{2}$, 64) $\rightarrow$ ($\frac{h}{4}$, $\frac{w}{4}$, 128) & CONV-(N128, K4x4, S2, P1), IN, ELU \\
		& ($\frac{h}{4}$, $\frac{w}{4}$, 128) $\rightarrow$ ($\frac{h}{8}$, $\frac{w}{8}$, 256) & CONV-(N256, K4x4, S2, P1), IN, ELU \\ 
		& ($\frac{h}{8}$, $\frac{w}{8}$, 256) $\rightarrow$ ($\frac{h}{16}$, $\frac{w}{16}$, 512) & CONV-(N512, K4x4, S2, P1), IN, ELU \\
		\midrule
		\multirow{10}{*}{$E$} & ($\frac{h}{16}$, $\frac{w}{16}$, 512 + 512) $\rightarrow$ ($\frac{h}{16}$, $\frac{w}{16}$, 1024) & CONCAT($E_{im}$, $E_{se}$)\\ 
		& ($\frac{h}{16}$, $\frac{w}{16}$, 1024) $\rightarrow$ ($\frac{h}{16}$, $\frac{w}{16}$, 1024) & Residual Block: CONV-(N1024, K3x3, S1, P1, D2), IN, ELU \\
		& ($\frac{h}{16}$, $\frac{w}{16}$, 1024) $\rightarrow$ ($\frac{h}{16}$, $\frac{w}{16}$, 1024) & Residual Block: CONV-(N1024, K3x3, S1, P1, D2), IN, ELU \\
		& ($\frac{h}{16}$, $\frac{w}{16}$, 1024) $\rightarrow$ ($\frac{h}{16}$, $\frac{w}{16}$, 1024) & Residual Block: CONV-(N1024, K3x3, S1, P1, D2), IN, ELU \\
		& ($\frac{h}{16}$, $\frac{w}{16}$, 1024) $\rightarrow$ ($\frac{h}{16}$, $\frac{w}{16}$, 1024) & Residual Block: CONV-(N1024, K3x3, S1, P1, D4), IN, ELU \\
		& ($\frac{h}{16}$, $\frac{w}{16}$, 1024) $\rightarrow$ ($\frac{h}{16}$, $\frac{w}{16}$, 1024) & Residual Block: CONV-(N1024, K3x3, S1, P1, D4), IN, ELU \\
		& ($\frac{h}{16}$, $\frac{w}{16}$, 1024) $\rightarrow$ ($\frac{h}{16}$, $\frac{w}{16}$, 1024) & Residual Block: CONV-(N1024, K3x3, S1, P1, D4), IN, ELU \\
		& ($\frac{h}{16}$, $\frac{w}{16}$, 1024) $\rightarrow$ ($\frac{h}{16}$, $\frac{w}{16}$, 1024) & Residual Block: CONV-(N1024, K3x3, S1, P1, D8), IN, ELU \\
		& ($\frac{h}{16}$, $\frac{w}{16}$, 1024) $\rightarrow$ ($\frac{h}{16}$, $\frac{w}{16}$, 1024) & Residual Block: CONV-(N1024, K3x3, S1, P1, D8), IN, ELU \\
		& ($\frac{h}{16}$, $\frac{w}{16}$, 1024) $\rightarrow$ ($\frac{h}{16}$, $\frac{w}{16}$, 1024) & Residual Block: CONV-(N1024, K3x3, S1, P1, D8), IN, ELU \\
		\midrule
		\multirow{6}{*}{$G$} & ($\frac{h}{16}$, $\frac{w}{16}$, 1024) $\rightarrow$ ($\frac{h}{8}$, $\frac{h}{8}$, 512) & Decoder Block \\
		& ($\frac{h}{8}$, $\frac{w}{8}$, 512) $\rightarrow$ ($\frac{h}{4}$, $\frac{h}{4}$, 256) & Decoder Block \\
		& ($\frac{w}{16}$, $\frac{w}{16}$, 256) $\rightarrow$ ($\frac{h}{2}$, $\frac{w}{2}$, 128) & Decoder Block \\
		& ($\frac{h}{2}$, $\frac{w}{2}$, 128) $\rightarrow$ ($h$, $w$, 64) & Decoder Block \\
		& ($h$, $w$, 64) $\rightarrow$ ($h$, $w$, 32) & SPADEResBlock \\
		& ($h$, $w$, 32) $\rightarrow$ ($h$, $w$, 3) & CONV-(N32, K7x7, S1, P3, D8), Tanh \\
		
		\midrule
		\multirow{4}{*}{$D_g$} & ($h$, $w$, 3 + $\mathcal{C}$) $\rightarrow$ ($\frac{h}{2}$, $\frac{w}{2}$, 64) & CONV-(N64, K4x4, S2, P1), Leaky ReLU, SN \\
		& ($\frac{h}{2}$, $\frac{w}{2}$, 64) $\rightarrow$ ($\frac{h}{4}$, $\frac{w}{4}$, 128) & CONV-(N128, K4x4, S2, P1), IN, Leaky ReLU, SN \\
		& ($\frac{h}{4}$, $\frac{w}{4}$, 128) $\rightarrow$ ($\frac{h}{8}$, $\frac{w}{8}$, 256) & CONV-(N256, K4x4, S2, P1), IN, Leaky ReLU \\
		& ($\frac{h}{8}$, $\frac{w}{8}$, 256) $\rightarrow$ ($\frac{h}{8}$, $\frac{w}{8}$, 512) & CONV-(N512, K4x4, S1, P1), IN, Leaky ReLU \\
		& ($\frac{h}{8}$, $\frac{w}{8}$, 512) $\rightarrow$ ($\frac{h}{8}$, $\frac{w}{8}$, 1) & CONV-(N1, K4x4, S1, P1) \\
		\bottomrule
	\end{tabular}
	\vspace{1mm}
	\caption{Network architecture. We use the following notation: $\bm{m}$: the dimension of the mask, $\mathcal{C}$: the semantic classes, $\mathbf{\hat{m}}_{s}$: the dimension of the instance, K: kernel size, S: stride size, P: padding size, D: dilation factor, CONV: a convolutional layer, IN: Instance Normalization, SN: Spectral Normalization }
	\label{tab:architecture_1}
\end{table*}

\section{Dataset details}
\noindent\textbf{Cityscapes.} The dataset contains 5000 street level images divided between training, validation and testing sets. We used the 2975 images from the training set during the training and we tested our model on the 500 images from the validation set. The resolution of the images is 2048 $\times$ 1024. We resize each image to be 512 $\times$ 256 in order to maintain the scale and then we apply random cropping of size 256 $\times$ 256.
Cityscapes segmentation maps have 35 categories. We aggregated them in 17 categories inspired by the  literature~\cite{cordts2016cityscapes}. 

\vpara{Indian Driving.} It consists of 20000 images collecting driving sequences on Indian roads. As for the Cityscapes, we used the validation set for the evaluation phase. We removed the 720p images and resized the remaining (1080p) images to 512 $\times$ 288. Thus, we obtain a dataset of 11564 training images and 1538 evaluation images. Among these evaluation images, we randomly select 500 for the evaluation. Finally, we apply random cropping of size 256 $\times$ 256.
To be comparable with Cityscapes, we aggregated the 40 segmentation maps categories into 21 groups.

\section{Use cases}
\begin{figure*}[ht]
	\setlength{\tabcolsep}{1pt}
	\renewcommand{\arraystretch}{0.8}
	\newcommand{\sizea}{0.19\linewidth}
	\footnotesize
	\centering
	\begin{tabular}{c cccc c}
		& Input image & SPG-Net & RN & Our & Ground truth \\
		\multirow{2}{*}{\rotatebox{90}{\small Cityscapes}} &
		\includegraphics[width=\sizea]{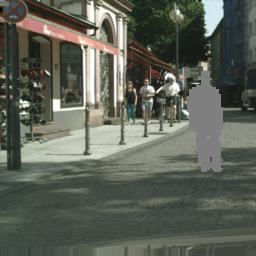}&
		\includegraphics[width=\sizea]{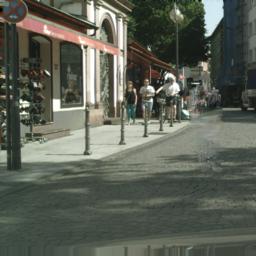}&
		\includegraphics[width=\sizea]{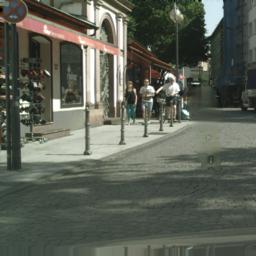}&
		\includegraphics[width=\sizea]{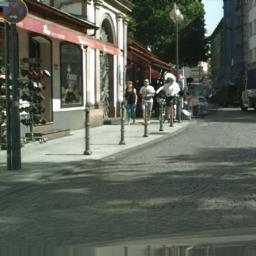}&
		\includegraphics[width=\sizea]{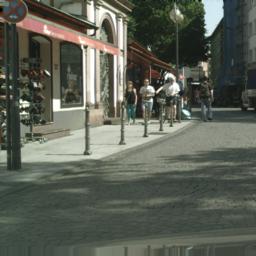}\\ 
		& \includegraphics[width=\sizea]{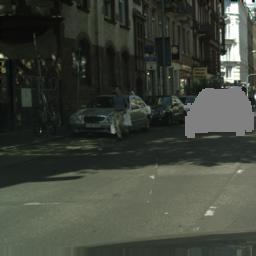}&
		\includegraphics[width=\sizea]{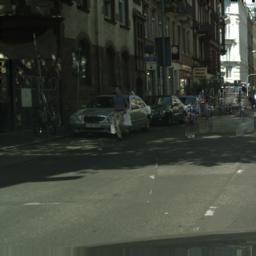}&
		\includegraphics[width=\sizea]{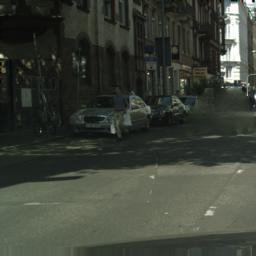}&
		\includegraphics[width=\sizea]{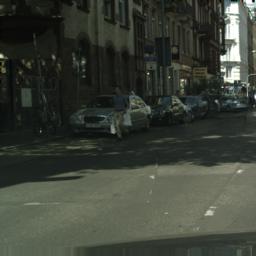}&
		\includegraphics[width=\sizea]{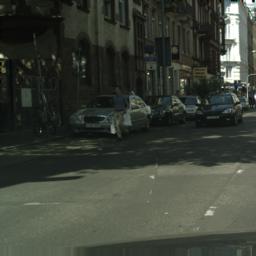}\\ 
		
		\multirow{2}{*}{\rotatebox{90}{\enspace \small Indian Driving}} & \includegraphics[width=\sizea]{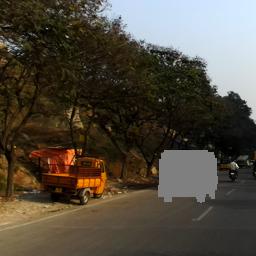}&
		\includegraphics[width=\sizea]{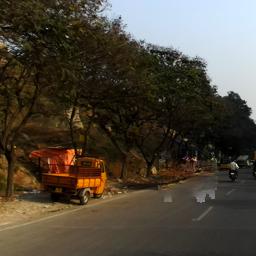}&
		\includegraphics[width=\sizea]{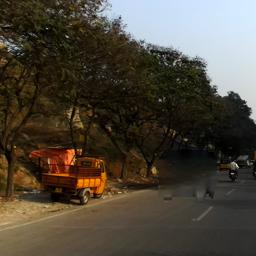}&
		\includegraphics[width=\sizea]{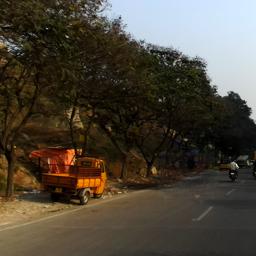}&
		\includegraphics[width=\sizea]{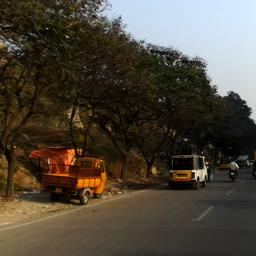}\\  
		& \includegraphics[width=\sizea]{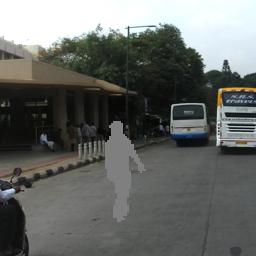}&
		\includegraphics[width=\sizea]{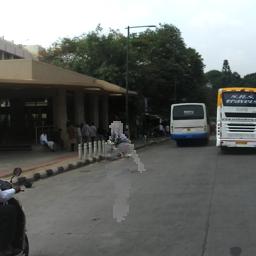}&
		\includegraphics[width=\sizea]{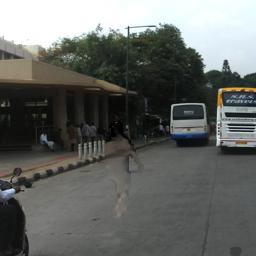}&
		\includegraphics[width=\sizea]{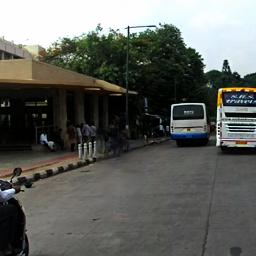}&
		\includegraphics[width=\sizea]{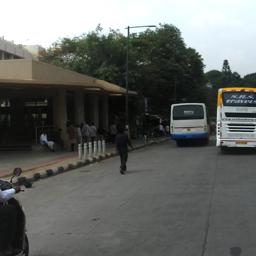}\\ 
	\end{tabular}
	\caption{Qualitative evaluation on the task of object precise removal. The first two rows show results on the Cityscape dataset while the last two rows show the Indian Driving results. We show two types of object removal: cars and pedestrians.}
	\label{Fig:preciseRemoval}
\end{figure*}

\begin{figure*}[ht]
	\setlength{\tabcolsep}{1pt}
	\renewcommand{\arraystretch}{0.8}
	\newcommand{\sizea}{0.12\linewidth}
	\footnotesize
	\centering
	\begin{tabular}{cc cccccc c}
		& Input image & Hong \textit{et al.}~\cite{hong2018learning} & Input object & SPG-Net* & RN* & Our & Our (segmentation) & Ground truth \\
		\multirow{2}{*}{\rotatebox{90}{\small Cityscapes}} &
		\includegraphics[width=\sizea]{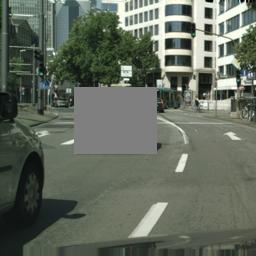} &
		\includegraphics[width=\sizea]{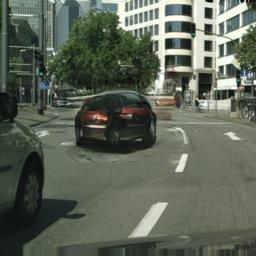} &
		\includegraphics[width=\sizea]{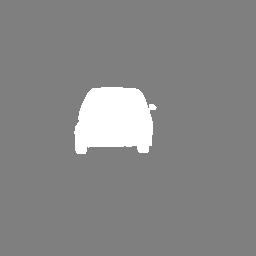} &
		\includegraphics[width=\sizea]{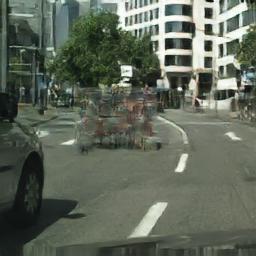} &
		\includegraphics[width=\sizea]{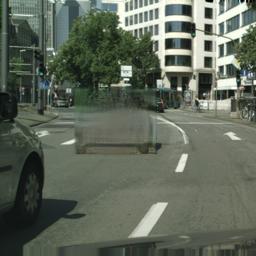} &
		\includegraphics[width=\sizea]{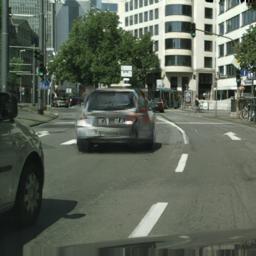} &
		\includegraphics[width=\sizea]{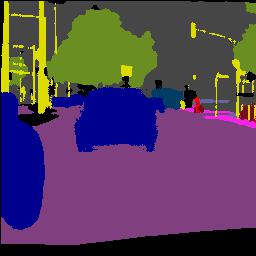} &
		\includegraphics[width=\sizea]{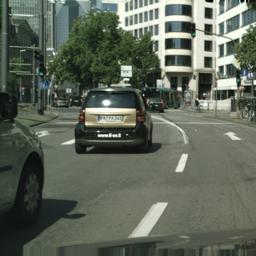} \\ 
		& \includegraphics[width=\sizea]{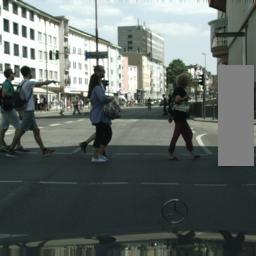} &
		\includegraphics[width=\sizea]{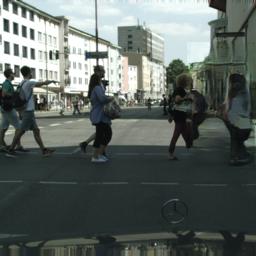} & 
		\includegraphics[width=\sizea]{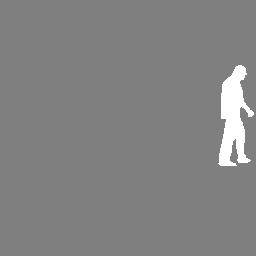} & 
		\includegraphics[width=\sizea]{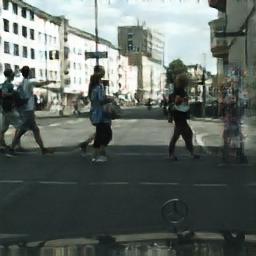} &
		\includegraphics[width=\sizea]{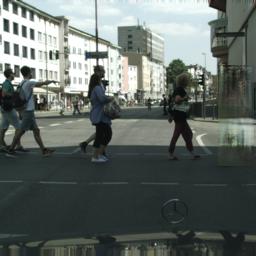} &
		\includegraphics[width=\sizea]{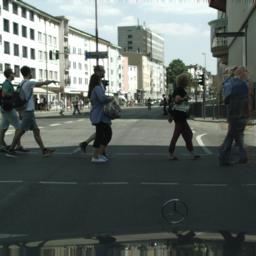} &
		\includegraphics[width=\sizea]{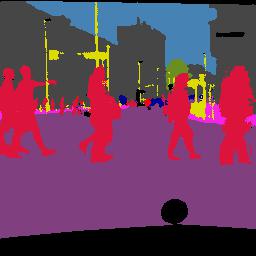} &
		\includegraphics[width=\sizea]{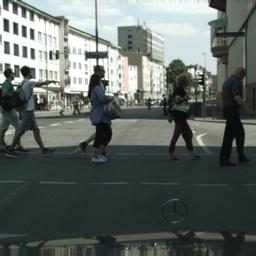} \\
		
		\multirow{2}{*}{\rotatebox{90}{\enspace \small Indian Driving}} &
		\includegraphics[width=\sizea]{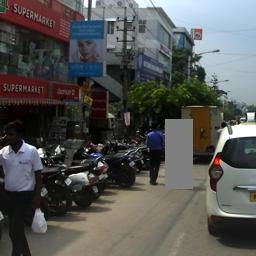} &
		\includegraphics[width=\sizea]{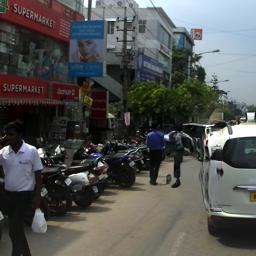} & 
		\includegraphics[width=\sizea]{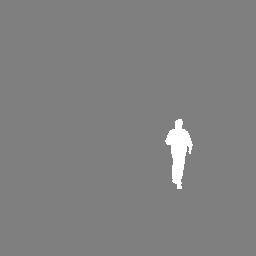} &
		\includegraphics[width=\sizea]{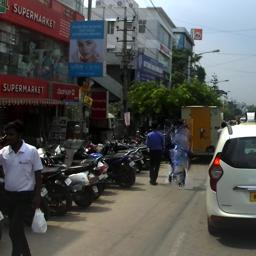} &
		\includegraphics[width=\sizea]{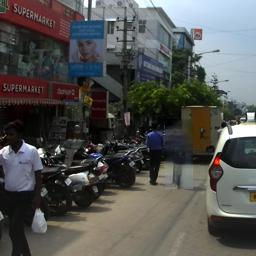} &
		\includegraphics[width=\sizea]{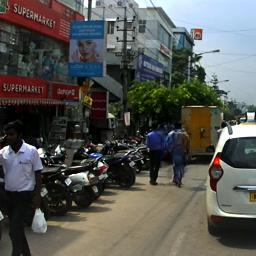} &
		\includegraphics[width=\sizea]{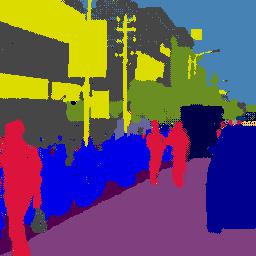} &
		\includegraphics[width=\sizea]{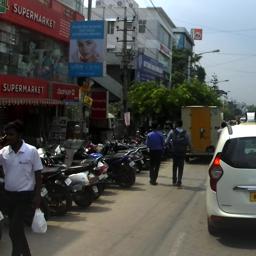} \\ 
		& \includegraphics[width=\sizea]{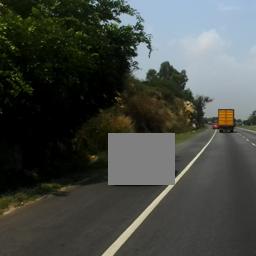} &
		\includegraphics[width=\sizea]{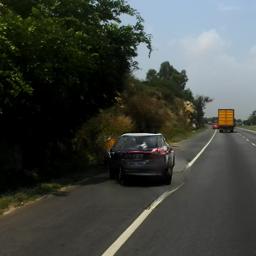} & 
		\includegraphics[width=\sizea]{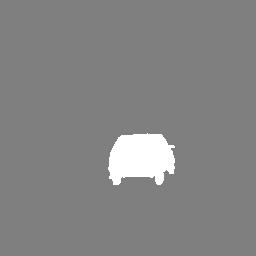} &  
		\includegraphics[width=\sizea]{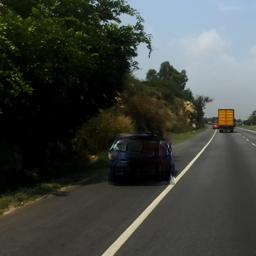} &
		\includegraphics[width=\sizea]{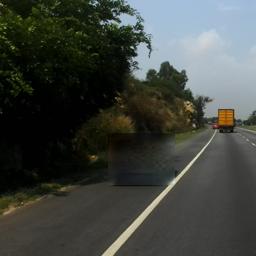} &
		\includegraphics[width=\sizea]{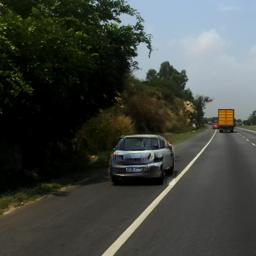} &
		\includegraphics[width=\sizea]{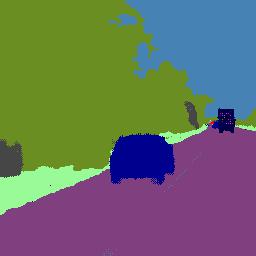} &
		\includegraphics[width=\sizea]{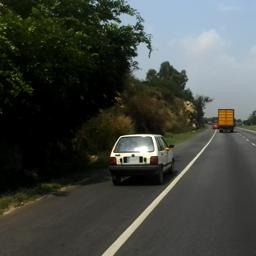} \\
	\end{tabular}
	\caption{Qualitative evaluation on the task of "mask insertion" object reconstruction. The first two rows show results on the Cityscape dataset while the last two rows show the Indian Driving results. We show two types of object insertion: cars and pedestrians.}
	\label{Fig:mask_insertion}
\end{figure*}

We test the network also on different uses cases listed in Figure 1 of the main paper, namely \emph{precise removal} and \emph{mask insertion}.
The former aims to remove the exact shape of the object by reconstructing background pixels, while the latter inserts a mask provided by the user as input.
We observe that our solution outperforms the state of the art both quantitatively (see \Cref{tab:results}) and qualitatively (see \Cref{Fig:mask_insertion} and \Cref{Fig:preciseRemoval}).

\begin{table}[!ht]
	\centering
	\caption{Quantitative results for our model and the baselines for the use cases of precise removal and maks insertion. We evaluate all the networks in two experimental settings and through image quality (PSNR and FID).}
	\label{tab:results}
	\begin{tabular}{@{}clc r rr r rr@{}}
		\midrule
		&\multirow{2}{*}{\textbf{Model}} &
		\multicolumn{2}{c}{\textbf{Cityscapes}} & \multicolumn{2}{c}{\textbf{Indian Driving}} \\
		\cmidrule(l{2pt}){3-4} \cmidrule(l{2pt}){5-6}
		&& PSNR$\uparrow$ & FID$\downarrow$ & PSNR$\uparrow$ & FID$\downarrow$ \\
		\midrule
		\multirow{4}{*}{\rotatebox{90}{Restore}} & Hong \emph{et al.}~\cite{hong2018learning} \\
		& SPG-Net~\cite{song2018spg} & 29.35 & 21.56 & 25.06 & 42.68\\
		& RN~\cite{yu2019region}  & 30.34 & 23.41 & 27.04 & 50.20 \\
		& Our proposal & \textbf{31.06} & \textbf{16.34} & \textbf{27.20} & \textbf{40.15}\\
		\midrule
		\multirow{4}{*}{\rotatebox{90}{Restore}} & Hong \emph{et al.}~\cite{hong2018learning} & 31.03 & 13.87 & 28.23 & 19.20\\
		& SPG-Net* & 31.36 & 17.96 & 27.44 & 26.96 \\
		& RN*  & 31.70 & 18.67 & 28.32 & 39.91 \\
		& Our proposal & \textbf{32.39} & \textbf{10.96}  & \textbf{28.85} & \textbf{19.02}  \\
		\bottomrule      
	\end{tabular}
\end{table}

\section{Additional results}
\begin{figure*}[ht]
	\setlength{\tabcolsep}{1pt}
	\renewcommand{\arraystretch}{0.8}
	\newcommand{\sizea}{0.15\linewidth}
	\footnotesize
	\centering
	\begin{tabular}{cccccc}
		Input image & Hong \textit{et al.}~\cite{hong2018learning} & SPG-Net  & RN & Our & GT \\
		\includegraphics[width=\sizea]{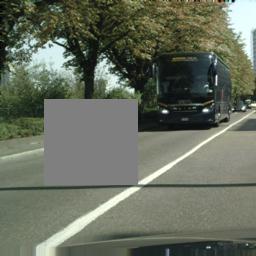} &
		\includegraphics[width=\sizea]{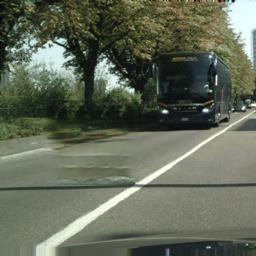} &
		\includegraphics[width=\sizea]{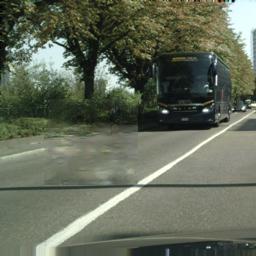} &
		\includegraphics[width=\sizea]{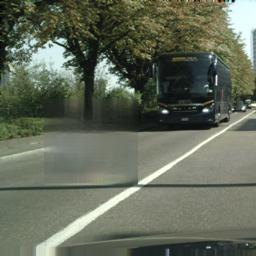} & \includegraphics[width=\sizea]{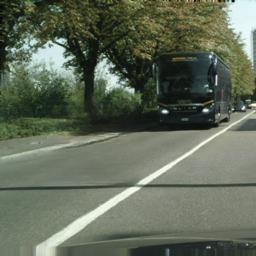} &
		\includegraphics[width=\sizea]{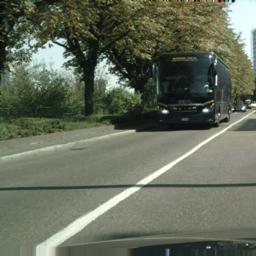}\\ 
		
		\includegraphics[width=\sizea]{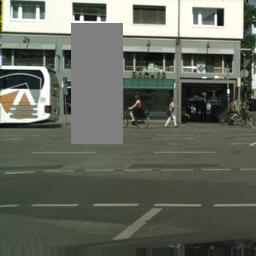} &
		\includegraphics[width=\sizea]{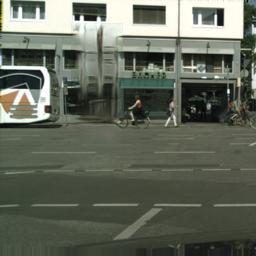} &
		\includegraphics[width=\sizea]{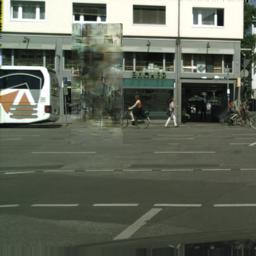} &
		
		\includegraphics[width=\sizea]{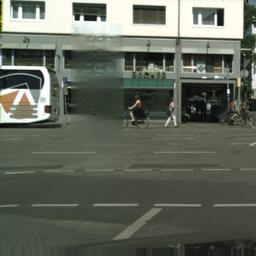} & \includegraphics[width=\sizea]{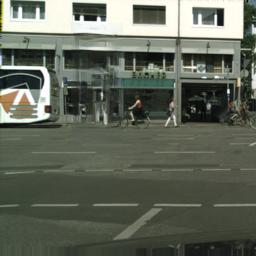} & \includegraphics[width=\sizea]{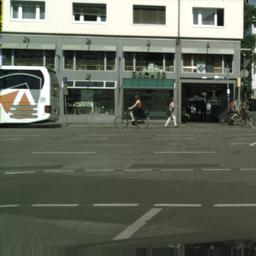} \\
		
		\includegraphics[width=\sizea]{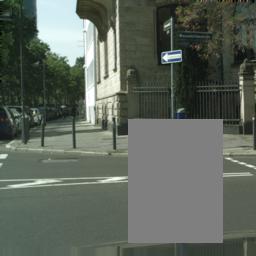} &
		\includegraphics[width=\sizea]{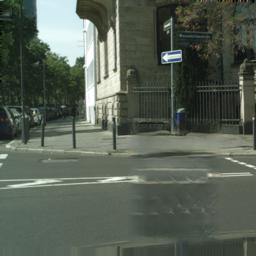} &
		\includegraphics[width=\sizea]{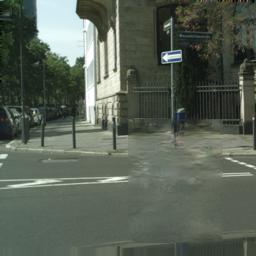} &
		
		\includegraphics[width=\sizea]{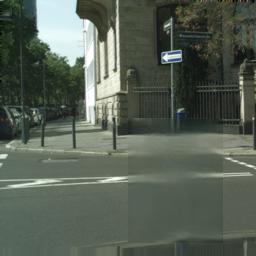} & \includegraphics[width=\sizea]{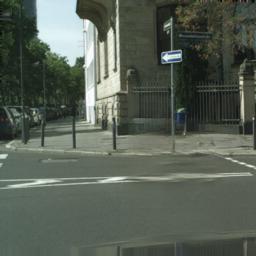} &
		\includegraphics[width=\sizea]{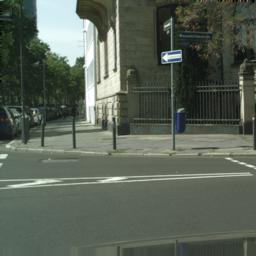}\\ 
		
		\includegraphics[width=\sizea]{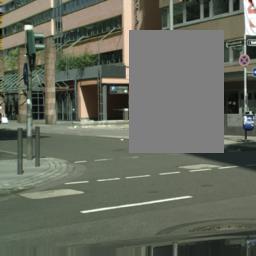} &
		\includegraphics[width=\sizea]{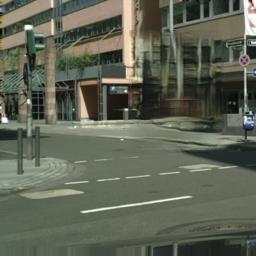} &
		\includegraphics[width=\sizea]{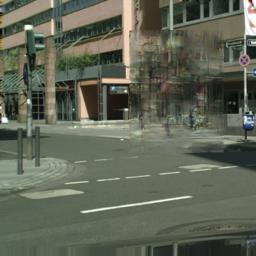} &
		
		\includegraphics[width=\sizea]{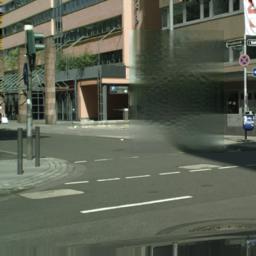} & \includegraphics[width=\sizea]{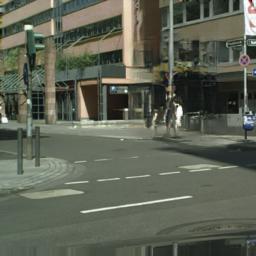} &
		\includegraphics[width=\sizea]{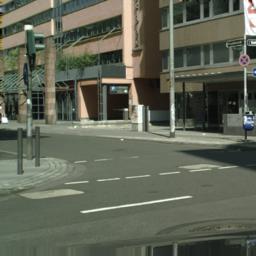}\\ 
		
		\includegraphics[width=\sizea]{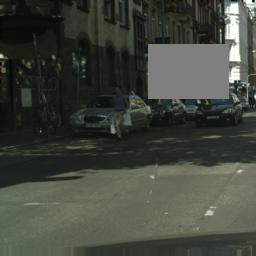} &
		\includegraphics[width=\sizea]{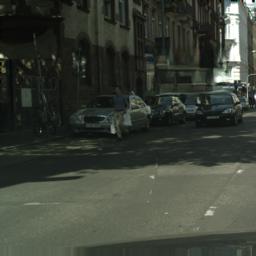} &
		\includegraphics[width=\sizea]{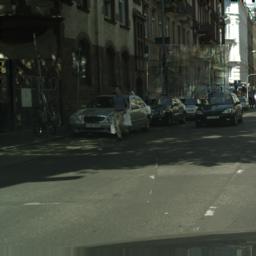} &
		
		\includegraphics[width=\sizea]{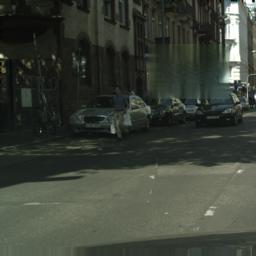} & \includegraphics[width=\sizea]{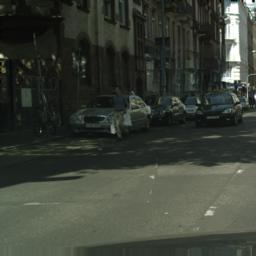} &
		\includegraphics[width=\sizea]{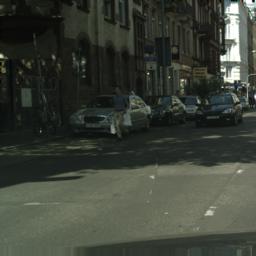}\\
	\end{tabular}
	\caption{Additional results on Cityscapes reconstruct experimental setting.}
	\label{Fig:cityscapes_no_cond}
\end{figure*}

\begin{figure*}[ht]
	\setlength{\tabcolsep}{1pt}
	\renewcommand{\arraystretch}{0.8}
	\newcommand{\sizea}{0.131\linewidth}
	\footnotesize
	\centering
	\begin{tabular}{ccccc c}
		Input image & Hong \textit{et al.}~\cite{hong2018learning} & SPG-Net & RN & Our & GT \\
		\includegraphics[width=\sizea]{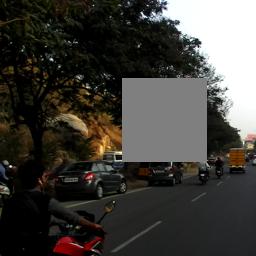} &
		\includegraphics[width=\sizea]{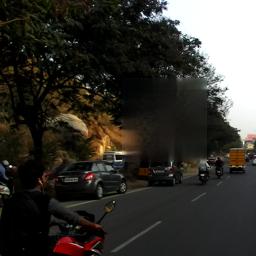} &
		\includegraphics[width=\sizea]{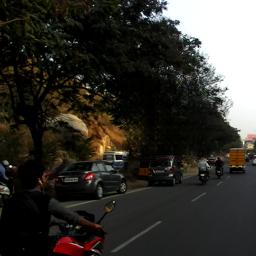} &
		\includegraphics[width=\sizea]{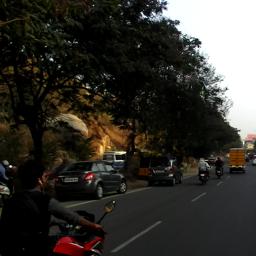} & 
		\includegraphics[width=\sizea]{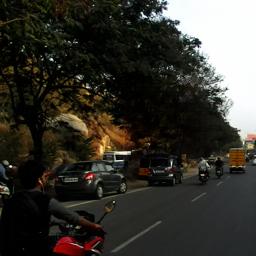} &
		\includegraphics[width=\sizea]{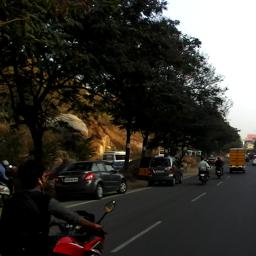}\\ 
		
		\includegraphics[width=\sizea]{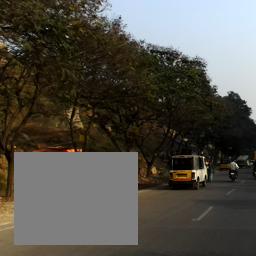} &
		\includegraphics[width=\sizea]{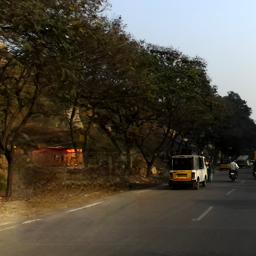} & 
		\includegraphics[width=\sizea]{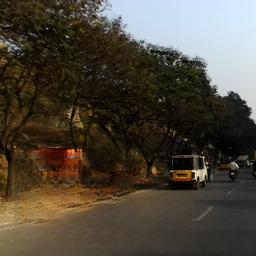} &
		
		\includegraphics[width=\sizea]{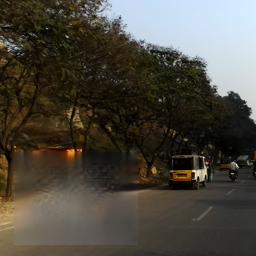} & \includegraphics[width=\sizea]{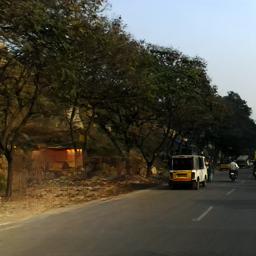} &
		\includegraphics[width=\sizea]{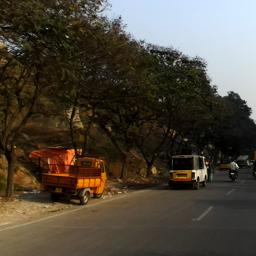}\\ 
		
		\includegraphics[width=\sizea]{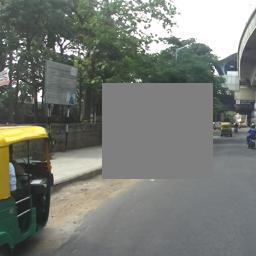} &
		\includegraphics[width=\sizea]{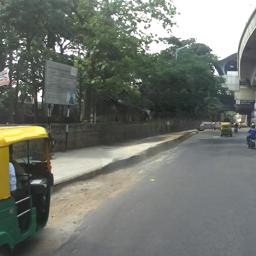} & 
		\includegraphics[width=\sizea]{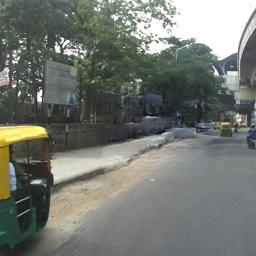} &
		
		\includegraphics[width=\sizea]{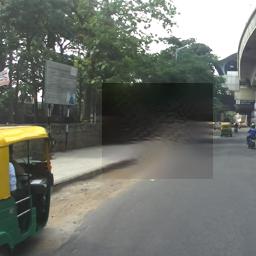} & \includegraphics[width=\sizea]{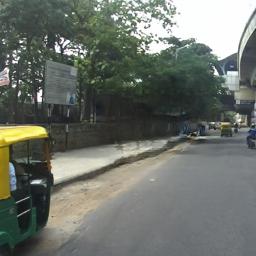} &
		\includegraphics[width=\sizea]{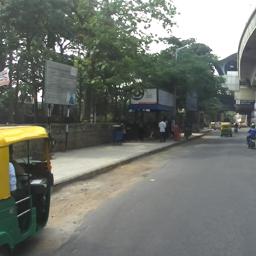}\\ 
		
		\includegraphics[width=\sizea]{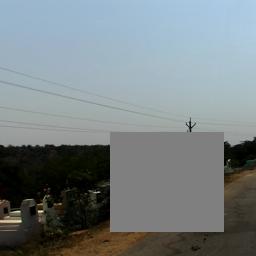} &
		\includegraphics[width=\sizea]{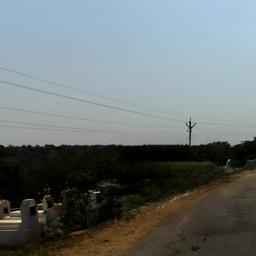} & 
		\includegraphics[width=\sizea]{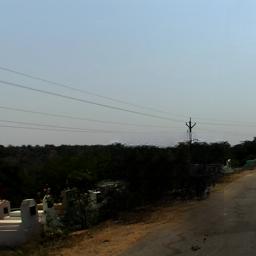} &
		
		\includegraphics[width=\sizea]{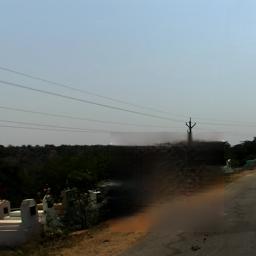} & \includegraphics[width=\sizea]{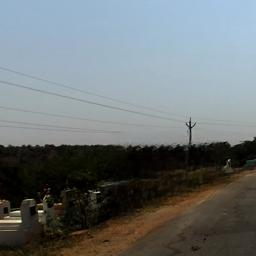} &
		\includegraphics[width=\sizea]{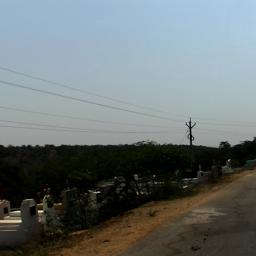}\\ 
		
		\includegraphics[width=\sizea]{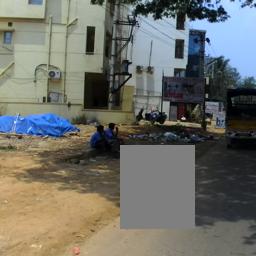} &
		\includegraphics[width=\sizea]{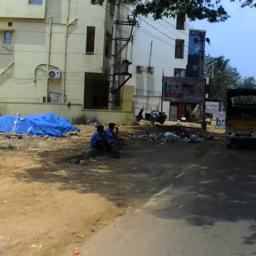} & 
		\includegraphics[width=\sizea]{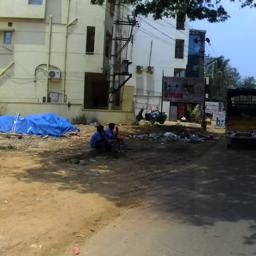} &
		
		\includegraphics[width=\sizea]{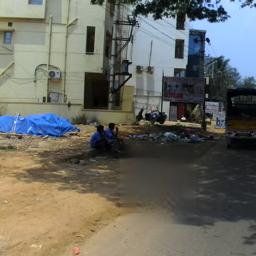} & \includegraphics[width=\sizea]{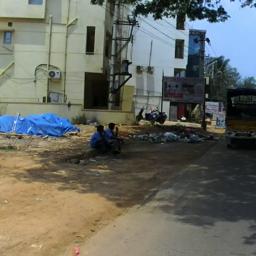} &
		\includegraphics[width=\sizea]{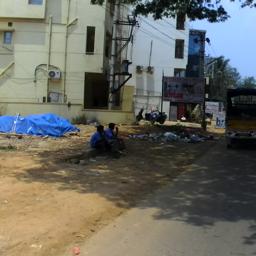}\\ 
	\end{tabular}
	\caption{Additional results on Indian Driving Dataset reconstruct experimental setting.}
	\label{Fig:vistas_no_cond}
\end{figure*}

\begin{figure*}[ht]
	\setlength{\tabcolsep}{1pt}
	\renewcommand{\arraystretch}{0.8}
	\newcommand{\sizea}{0.131\linewidth}
	\footnotesize
	\centering
	\begin{tabular}{cc ccccc c}
		& Input image & Hong \textit{et al.}~\cite{hong2018learning} & SPG-Net* & RN* & Our & Our (segmentation) & Ground truth \\
		\rotatebox[origin=c]{90}{\small Car} & 
		\raisebox{-0.5\height}{\includegraphics[width=\sizea]{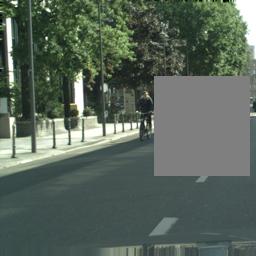}}&
		\raisebox{-0.5\height}{\includegraphics[width=\sizea]{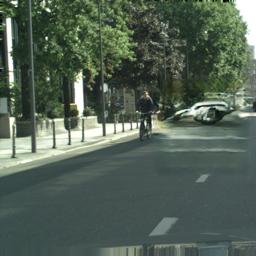}}&
		\raisebox{-0.5\height}{\includegraphics[width=\sizea]{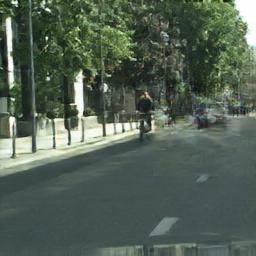}} &
		\raisebox{-0.5\height}{ \includegraphics[width=\sizea]{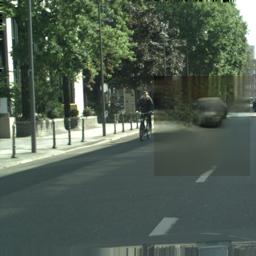}} &
		\raisebox{-0.5\height}{\includegraphics[width=\sizea]{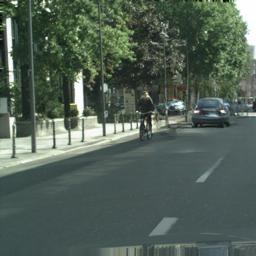}} &
		\raisebox{-0.5\height}{\includegraphics[width=\sizea]{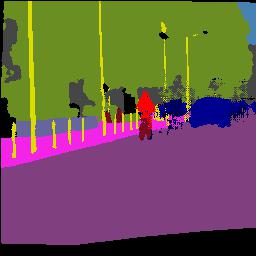}} &
		\raisebox{-0.5\height}{\includegraphics[width=\sizea]{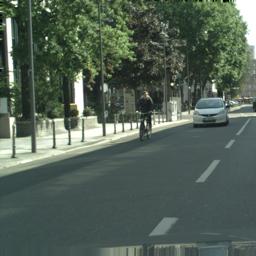}}\\

		\rotatebox[origin=c]{90}{\small Car} & 
		\raisebox{-0.5\height}{\includegraphics[width=\sizea]{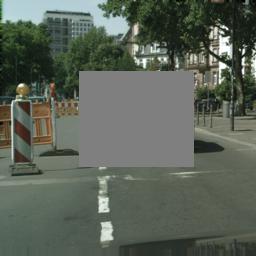}}&
		\raisebox{-0.5\height}{\includegraphics[width=\sizea]{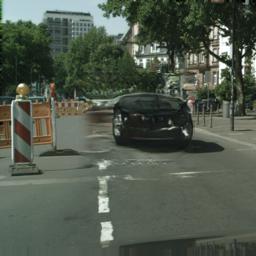}}& 
		\raisebox{-0.5\height}{\includegraphics[width=\sizea]{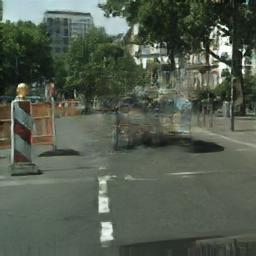}} &
		\raisebox{-0.5\height}{\includegraphics[width=\sizea]{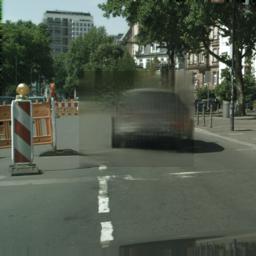}} &
		\raisebox{-0.5\height}{\includegraphics[width=\sizea]{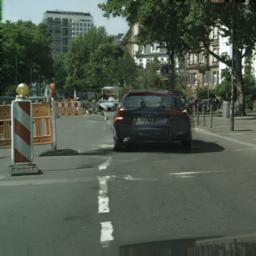}} &
		\raisebox{-0.5\height}{\includegraphics[width=\sizea]{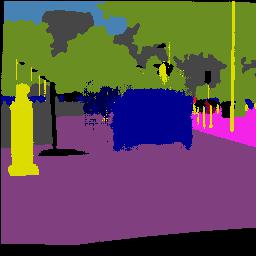}} &
		\raisebox{-0.5\height}{\includegraphics[width=\sizea]{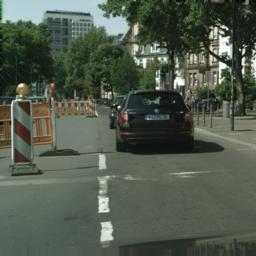}}\\

		\rotatebox[origin=c]{90}{\small Pedestrian} & 
		\raisebox{-0.5\height}{\includegraphics[width=\sizea]{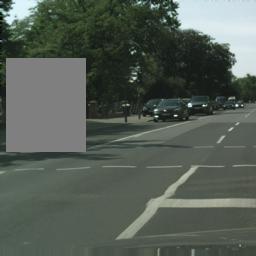}}&
		\raisebox{-0.5\height}{\includegraphics[width=\sizea]{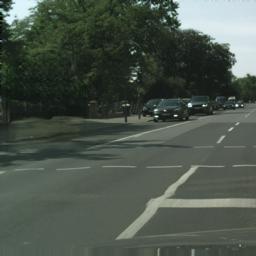}}&  
		\raisebox{-0.5\height}{\includegraphics[width=\sizea]{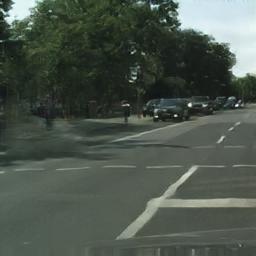}} &
		\raisebox{-0.5\height}{\includegraphics[width=\sizea]{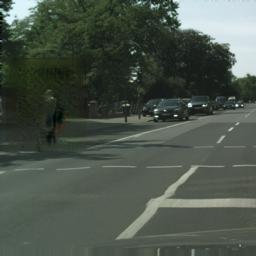}} &
		\raisebox{-0.5\height}{\includegraphics[width=\sizea]{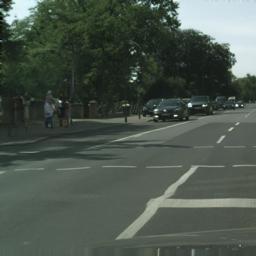}} &
		\raisebox{-0.5\height}{\includegraphics[width=\sizea]{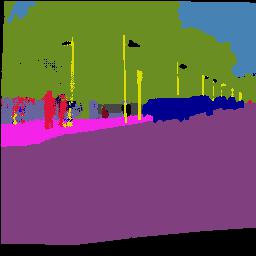}} &
		\raisebox{-0.5\height}{\includegraphics[width=\sizea]{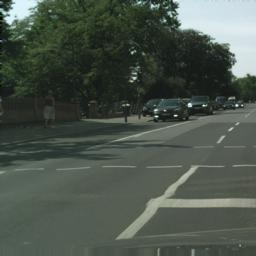}}\\

		\rotatebox[origin=c]{90}{\small Pedestrian} & 
		\raisebox{-0.5\height}{\includegraphics[width=\sizea]{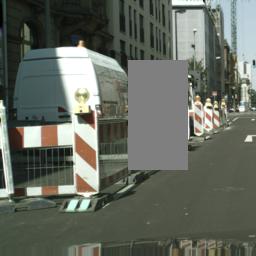}}&
		\raisebox{-0.5\height}{\includegraphics[width=\sizea]{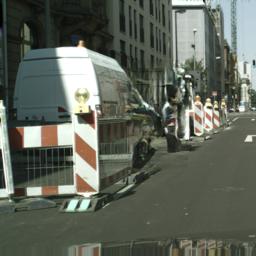}}&
		\raisebox{-0.5\height}{\includegraphics[width=\sizea]{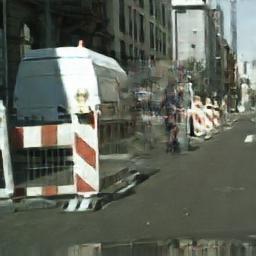}} &
		\raisebox{-0.5\height}{\includegraphics[width=\sizea]{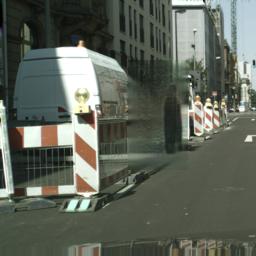}} &
		\raisebox{-0.5\height}{\includegraphics[width=\sizea]{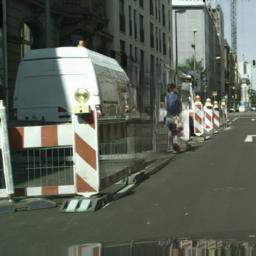}} &
		\raisebox{-0.5\height}{\includegraphics[width=\sizea]{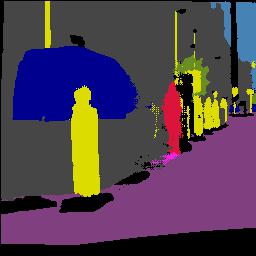}} &
		\raisebox{-0.5\height}{ \includegraphics[width=\sizea]{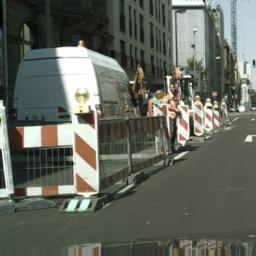}}\\

		\rotatebox[origin=c]{90}{\small Pedestrian} & 
		\raisebox{-0.5\height}{\includegraphics[width=\sizea]{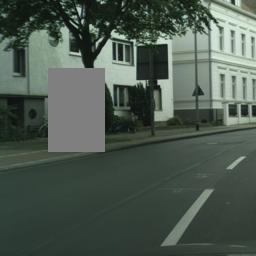}}&
		\raisebox{-0.5\height}{\includegraphics[width=\sizea]{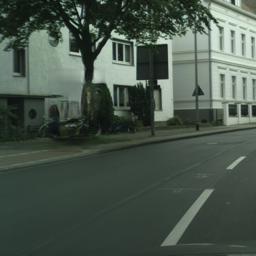}}&  
		\raisebox{-0.5\height}{\includegraphics[width=\sizea]{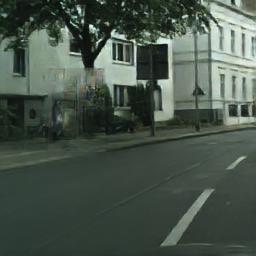}} &
		\raisebox{-0.5\height}{\includegraphics[width=\sizea]{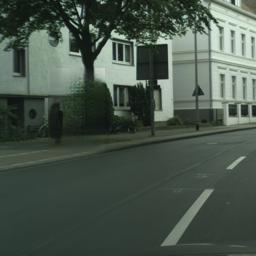}} &
		\raisebox{-0.5\height}{\includegraphics[width=\sizea]{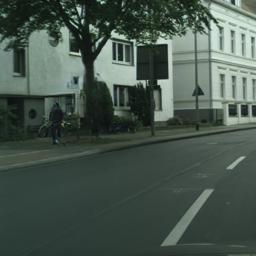}} &
		\raisebox{-0.5\height}{\includegraphics[width=\sizea]{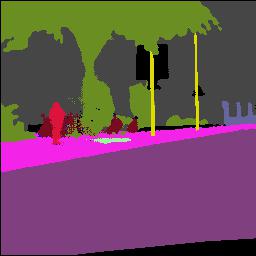}} &
		\raisebox{-0.5\height}{\includegraphics[width=\sizea]{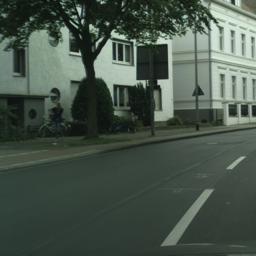}}\\ 
		
		\rotatebox[origin=c]{90}{\small Pedestrian} & 
		\raisebox{-0.5\height}{\includegraphics[width=\sizea]{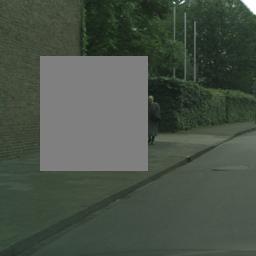}}&
		\raisebox{-0.5\height}{\includegraphics[width=\sizea]{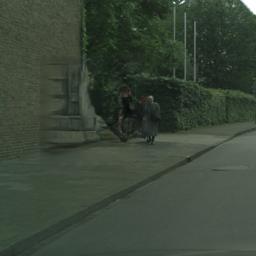}}& 
		\raisebox{-0.5\height}{\includegraphics[width=\sizea]{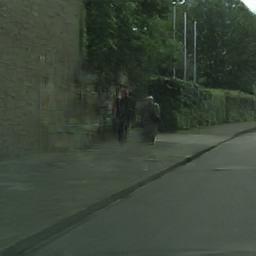}} &
		\raisebox{-0.5\height}{\includegraphics[width=\sizea]{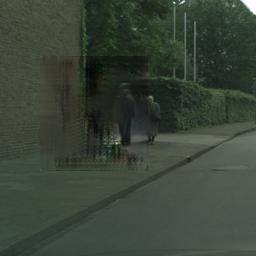}} &
		\raisebox{-0.5\height}{\includegraphics[width=\sizea]{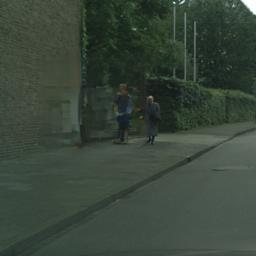}} &
		\raisebox{-0.5\height}{\includegraphics[width=\sizea]{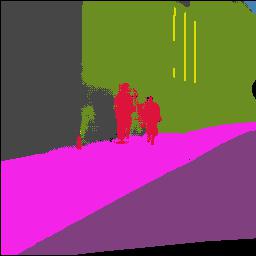}} &
		\raisebox{-0.5\height}{\includegraphics[width=\sizea]{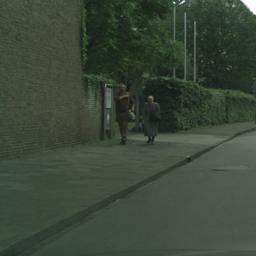}}\\ 
	\end{tabular}
	\caption{Additional results on Cityscapes Dataset insert \& reconstruct experimental setting.}
	\label{Fig:cityscapes_Cond}
\end{figure*}

\begin{figure*}[ht]
	\setlength{\tabcolsep}{1pt}
	\renewcommand{\arraystretch}{0.8}
	\newcommand{\sizea}{0.131\linewidth}
	\footnotesize
	\centering
	\begin{tabular}{cc ccccc c}
		& Input image & Hong \textit{et al.}~\cite{hong2018learning} & SPG-Net* & RN* & Our & Our (segmentation) & Ground truth \\
		\rotatebox[origin=c]{90}{\small Car} & 
		\raisebox{-0.5\height}{\includegraphics[width=\sizea]{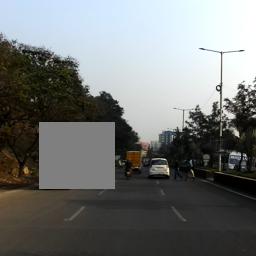}}&
		\raisebox{-0.5\height}{\includegraphics[width=\sizea]{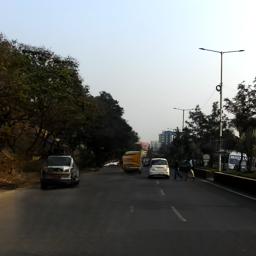}}&  
		\raisebox{-0.5\height}{\includegraphics[width=\sizea]{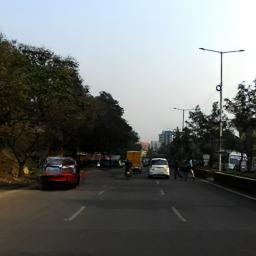}} &
		\raisebox{-0.5\height}{\includegraphics[width=\sizea]{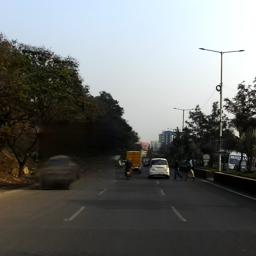}} &
		\raisebox{-0.5\height}{\includegraphics[width=\sizea]{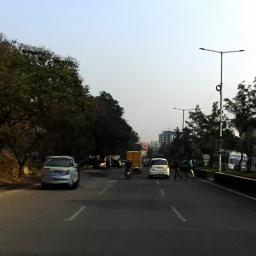}} &
		\raisebox{-0.5\height}{\includegraphics[width=\sizea]{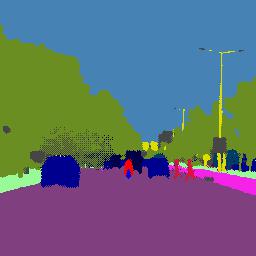}} &
		\raisebox{-0.5\height}{\includegraphics[width=\sizea]{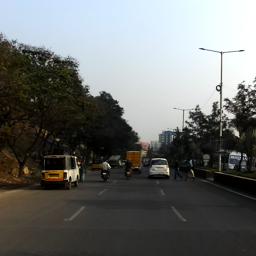}}\\ 
		
		\rotatebox[origin=c]{90}{\small Pedestrian} & 
		\raisebox{-0.5\height}{\includegraphics[width=\sizea]{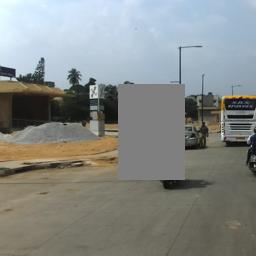}}&
		\raisebox{-0.5\height}{\includegraphics[width=\sizea]{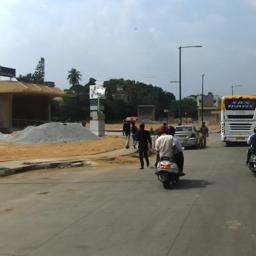}}&
		\raisebox{-0.5\height}{\includegraphics[width=\sizea]{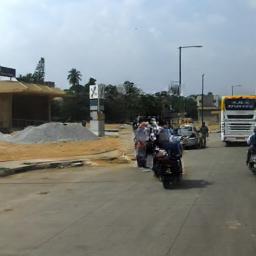}} &
		\raisebox{-0.5\height}{\includegraphics[width=\sizea]{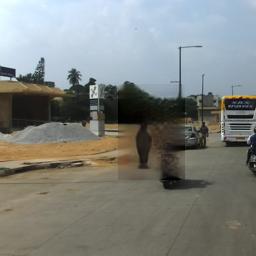}} &
		\raisebox{-0.5\height}{\includegraphics[width=\sizea]{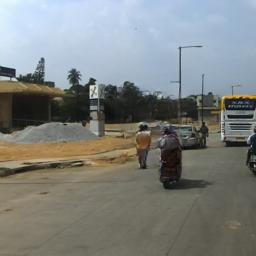}} &
		\raisebox{-0.5\height}{\includegraphics[width=\sizea]{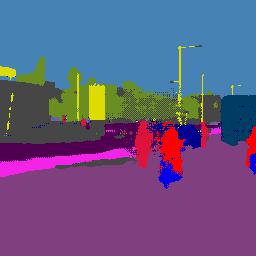}} &
		\raisebox{-0.5\height}{\includegraphics[width=\sizea]{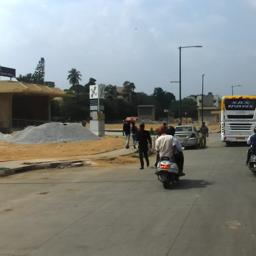}}\\
		
		\rotatebox[origin=c]{90}{\small Pedestrian} & 
		\raisebox{-0.5\height}{\includegraphics[width=\sizea]{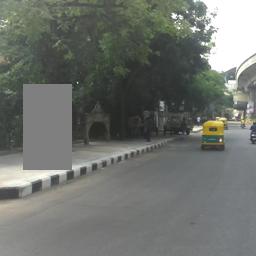}}&
		\raisebox{-0.5\height}{\includegraphics[width=\sizea]{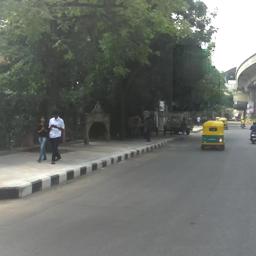}}&  
		\raisebox{-0.5\height}{\includegraphics[width=\sizea]{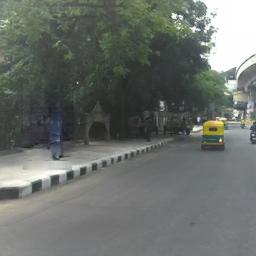}} &
		\raisebox{-0.5\height}{\includegraphics[width=\sizea]{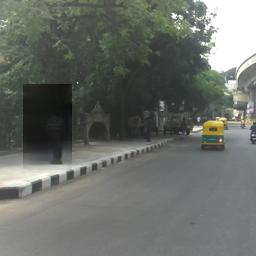}} &
		\raisebox{-0.5\height}{\includegraphics[width=\sizea]{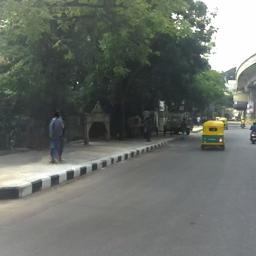}} &
		\raisebox{-0.5\height}{\includegraphics[width=\sizea]{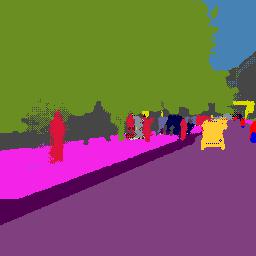}} &
		\raisebox{-0.5\height}{\includegraphics[width=\sizea]{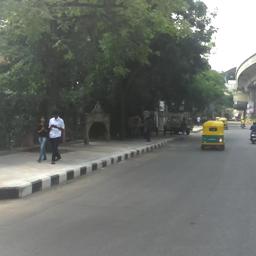}}\\
		\rotatebox[origin=c]{90}{\small Car} &\raisebox{-0.5\height}{\includegraphics[width=\sizea]{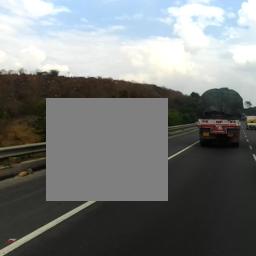}}&
		\raisebox{-0.5\height}{\includegraphics[width=\sizea]{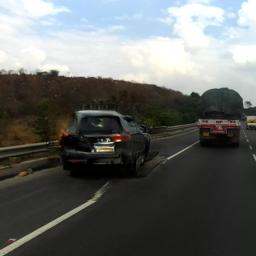}}&  
		\raisebox{-0.5\height}{\includegraphics[width=\sizea]{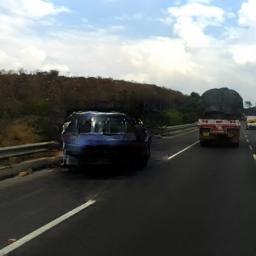}} &
		\raisebox{-0.5\height}{\includegraphics[width=\sizea]{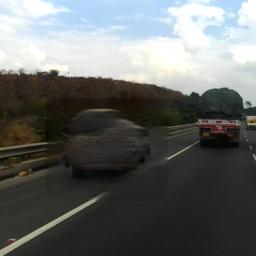}} &
		\raisebox{-0.5\height}{\includegraphics[width=\sizea]{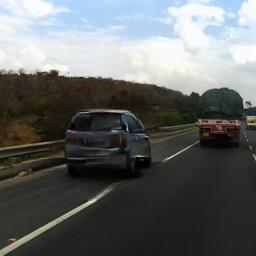}} &
		\raisebox{-0.5\height}{\includegraphics[width=\sizea]{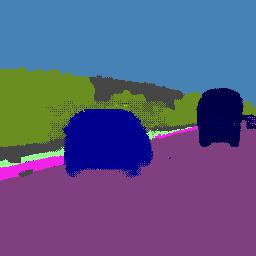}} &
		\raisebox{-0.5\height}{\includegraphics[width=\sizea]{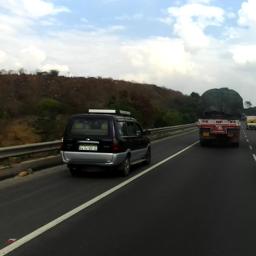}}\\
		\rotatebox[origin=c]{90}{\small Car} & \raisebox{-0.5\height}{\includegraphics[width=\sizea]{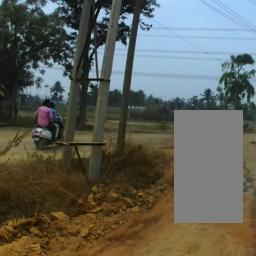}}&
		\raisebox{-0.5\height}{\includegraphics[width=\sizea]{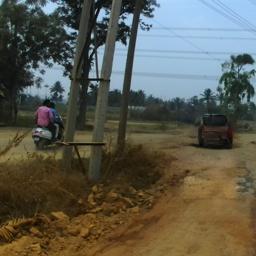}}&
		\raisebox{-0.5\height}{\includegraphics[width=\sizea]{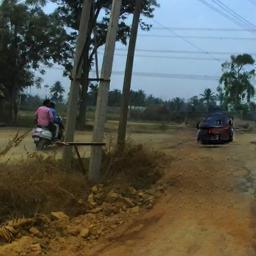}} &
		\raisebox{-0.5\height}{\includegraphics[width=\sizea]{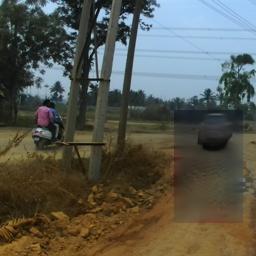}} &
		\raisebox{-0.5\height}{\includegraphics[width=\sizea]{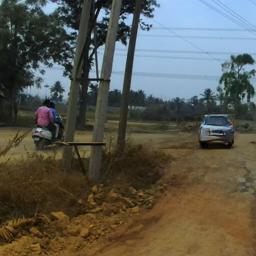}} &
		\raisebox{-0.5\height}{\includegraphics[width=\sizea]{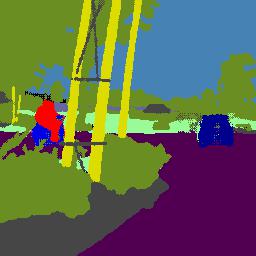}} &
		\raisebox{-0.5\height}{\includegraphics[width=\sizea]{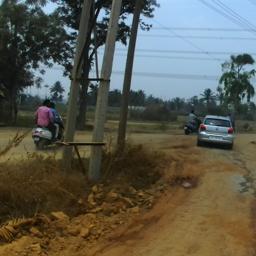}}\\
	\end{tabular}
	\caption{Additional results on Indian Driving Dataset insert \& reconstruct experimental setting.}
	\label{Fig:vistas_Cond}
\end{figure*}

We show additional result for each dataset and for each experimental settings. In particular, \Cref{Fig:cityscapes_no_cond} and \Cref{Fig:vistas_no_cond} show additional results for both the the Cityscapes and the Indian Driving datasets for the \emph{reconstruct} experimental settings. While, \Cref{Fig:cityscapes_Cond} and \Cref{Fig:vistas_Cond} for the \emph{insert \& reconstruct} experimental settings.

\section{Edges contours}
\begin{figure*}[ht]
	\setlength{\tabcolsep}{1pt}
	\renewcommand{\arraystretch}{0.8}
	\newcommand{\sizea}{0.21\linewidth}
	\footnotesize
	\centering
	\begin{tabular}{ccc}
		GT & Input image & Edges \\
		\includegraphics[width=\sizea]{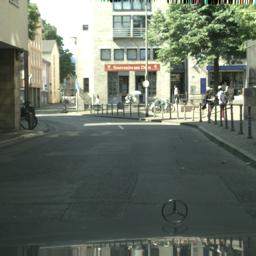} &
		\includegraphics[width=\sizea]{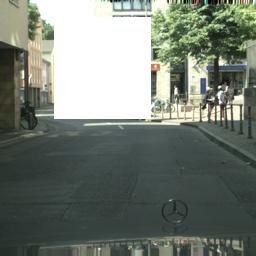} &
		\includegraphics[width=\sizea]{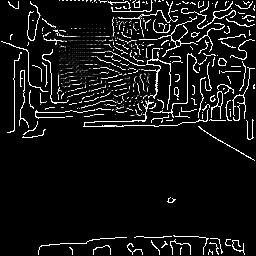}\\
		\includegraphics[width=\sizea]{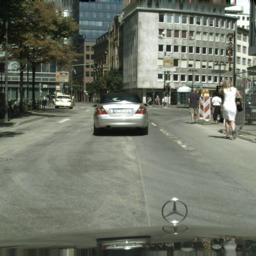} &
		\includegraphics[width=\sizea]{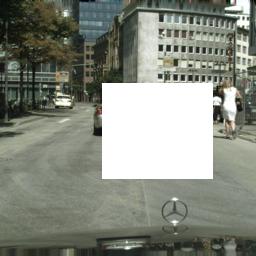} &
		\includegraphics[width=\sizea]{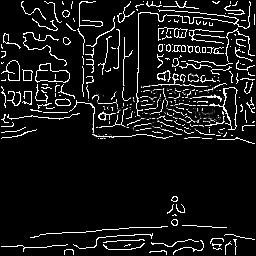} \\
	\end{tabular}
	\caption{Results of \cite{nazeri2019edgeconnect} on the Cityscapes dataset.}
	\label{Fig:nazeri}
\end{figure*}

In complex scenes, conventional inpainting methods might generate unsatisfactory results. For example, edge generators might underperform when the missing areas are either big or when they contain multiple objects and semantics. 
Thus, we show EdgeConnect~\cite{nazeri2019edgeconnect} results, which is a popular inpaiting method that exploits edges to inpaint missing regions.
\Cref{Fig:nazeri} shows the results of the edge inpainting in Cityscapes. As can be seen, when the portion of image to inpaint is large, the network is not capable of generate plausible edges. Indeed, in the first row it fails to generate the border of the building. While in the second row it generates a huge quantity of lines that do not fit the context of the object. Generated edges in complex scenes are often not reliable.

\section{Additional ablations}
As additional tests, we also ablate the different components of our decoder block. In our decoder, we removed the learned convolution for the residual skip connection, and replaced the Batch Normalization ~\cite{ioffe2015batch} with Instance Normalization~\cite{ulyanov2016instance}.
\Cref{tab:ablation} also shows that our modifications to the original SPADE~\cite{park2019semantic} are effective. Moreover, they allow using this block when the batch size is small (or even one).

\begin{figure}[ht]
	\setlength{\tabcolsep}{1pt}
	\renewcommand{\arraystretch}{0.8}
	\newcommand{\sizea}{0.21\linewidth}
	\footnotesize
	\centering
	\begin{tabular}{c ccc c}
		Input image & Our & Our w/o $\mathcal{L}_{\textrm{style}}$ \\
		\includegraphics[width=\sizea]{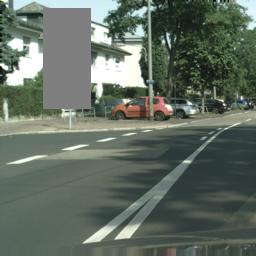} &
		\includegraphics[width=\sizea]{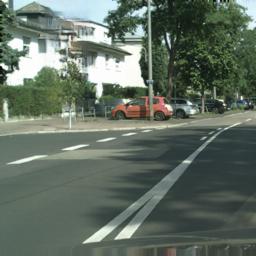} &
		\includegraphics[width=\sizea]{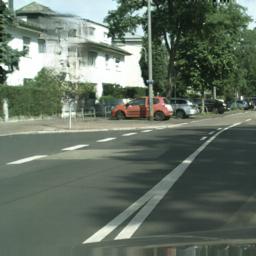} \\
	\end{tabular}
	\caption{Qualitative evaluation on the ablation study with and without $\mathcal{L}_{\textrm{style}}$. Removing $\mathcal{L}_{\textrm{style}}$ the network generate "checkerboard" artifacts. Zoom in for better details.}
	\label{Fig:ablation}
\end{figure}

In \Cref{Fig:ablation} we instead show the qualitative evaluation on ablation study with and without $\mathcal{L}_{\textrm{style}}$. The figure shows that removing $\mathcal{L}_{\textrm{style}}$ the network generate "checkerboard" artifacts.

\begin{table}[!h]
	\centering
	\caption{Ablation study of the components of the SPADE block.}
	\label{tab:ablation}
	\begin{tabular}{@{}l rrrr@{}}
		\toprule
		\textbf{Model} & \textbf{PSNR$\uparrow$} & \textbf{FID$\downarrow$}\\
		\midrule
		Our proposal (A) &  \textbf{32.96} & \textbf{5.05}\\
		(A) w/o SPADE & 32.57 & 5.56\\ 
		(A) w Sync Batch Normalization & 32.05 & 5.59\\
		(A) w learned skip connection & 32.76 & 5.49\\
		\bottomrule      
	\end{tabular}
\end{table}

\end{document}